\documentclass[a4paper,10pt]{RoManSy}

\usepackage[reqno]{amsmath}% AMS-LaTeX-first 
\usepackage{amssymb}
\usepackage{amsthm}  
\theoremstyle{plain}

\theoremstyle{definition}

\usepackage{psfrag}
\usepackage{epsfig}
\usepackage{graphicx}

\newcommand{\be}{\begin{equation}}
\newcommand{\ee}{\end{equation}}
\newcommand{\beq}{\begin{equation}}
\newcommand{\eeq}{\end{equation}}
\newcommand{\bed}{\begin{displaymath}}
\newcommand{\eed}{\end{displaymath}}
\newcommand{\beqa}{\begin{eqnarray}}
\newcommand{\eeqa}{\end{eqnarray}}
\newcommand{\beqann}{\begin{eqnarray*}}
\newcommand{\eeqann}{\end{eqnarray*}}
\newcommand{\bseq}{\begin{subequations}}
\newcommand{\eseq}{\end{subequations}}

\newcommand{\ba}{\begin{array}}
\newcommand{\ea}{\end{array}}

\newcommand{\negr}[1]{{\bf {#1}}}

\newtheorem{Def}{Definition}

\begin{document}

\title{\bf{Joint space and workspace analysis of a two-DOF closed-chain manipulator}}
\author{\normalsize{Damien Chablat} \\
  \normalsize{Institut de Recherche en Communications et Cybern\'etique de Nantes,} \\
  \normalsize{UMR CBRS 6597, 1 rue de la No$\ddot{e}$, Nantes, France} \\
  \normalsize{{damien.chablat@irccyn.ec-nantes.fr}}}
\maketitle

%%%%%%%%%%%%%%%%%%%%%%%%%%%%%%%%%%%%%%%%%%%%%%%%%%%%%%%%%%%%%%%%%%%%%%%%%%%%%%%%
\abstract{{\bf Abstract: } The aim of this paper is to compute of the generalized aspects, i.e. the maximal singularity-free domains in the Cartesian product of the joint space and workspace, for a planar parallel mechanism in using quadtree model and interval analysis based method. The parallel mechanisms can admit several solutions to the inverses and the direct kinematic models. These singular configurations divide the joint space and the workspace in several not connected domains.  To compute this domains, the quadtree model can be made by using a discretization of the space. Unfortunately, with this method, some singular configurations cannot be detected as a single point in the joint space. The interval analysis based method allow us to assure that all the singularities are found and to reduce the computing times. This approach is tested on a simple planar parallel mechanism with two degrees of freedom. }
%%%%%%%%%%%%%%%%%%%%%%%%%%%%%%%%%%%%%%%%%%%%%%%%%%%%%%%%%%%%%%%%%%%%%%%%%%%%%%%%
\section{Introduction}
%%%%%%%%%%%%%%%%%%%%%%%%%%%%%%%%%%%%%%%%%%%%%%%%%%%%%%%%%%%%%%%%%%%%%%%%%%%%%%%%
The kinematic design of parallel mechanism has drawn the interest of several researchers. The workspace is usually considered as a relevant design criterion \cite{Merlet:96,Clavel:88,Gosselin:91}. Parallel singularities \cite{Wenger:1997} occur in the workspace where the moving platform cannot resist any effort. They are very undesirable and generally eliminated by design. Serial singularities \cite{Chablat:1998} occur if the mechanism admit several solutions to the inverse kinematic model. To cope with the existence of multiple inverse kinematic solutions in {\em serial} manipulators, the notion of aspects was introduced in
\cite{Borrel:1986}. The aspects equal the maximal singularity-free domains in the joint space. For usual industrial serial manipulators, the aspects were found to be the maximal sets in the joint space where there is only one inverse kinematic solution.
\par
A definition of the notion of aspect was given by \cite{Wenger:1997} for parallel manipulators with only one inverse kinematic solution and was extended by \cite{Chablat:1998} for parallel manipulators with several solutions to the inverse and direct kinematic problem. These aspects were defined as the maximal singularity-free domains in the Cartesian product of the joint space and the workspace. To compute the aspects, we can used discretization methods. However, we cannot find any singularity in particular if the singularity is a point. Interval based analysis method was implemented by \cite{Merlet:2000,Chablat:2004} in {ALIAS} to compute the workspace of parallel mechanism. However, the result is a set of boxes in which is not easy to separate the maximum singularity free regions of the workspace and the computational times is difficult to estimate. 

This article introduces an algorithm based on the tree-like structure and the interval analysis based method which takes the advantages of the interval analysis based method and the simplicity of tree-like structures.
 %%%%%%%%%%%%%%%%%%%%%%%%%%%%%%%%%%%%%%%%%%%%%%%%%%%%%%%%%%%%%%%%%%%%%%%%%%%%%%%%
\section{Algorithm}
%%%%%%%%%%%%%%%%%%%%%%%%%%%%%%%%%%%%%%%%%%%%%%%%%%%%%%%%%%%%%%%%%%%%%%%%%%%%%%%%
The aim of this section is to define an algorithm able to compute either the joint space or the workspace of parallel mechanism. This is done using the structure of a quadtree model and the interval based method. This algorithm will be illustrated by a planar parallel mechanism in section \ref{example} but can easily be extended to mechanism with three degrees of freedom in using octree model. Unlike numerical computing methods, such a method allows to prove formally that there is no singular configuration in the final result.
%%%%%%%%%%%%%%%%%%%%%%%%%%%%%%%%%%%%%%%%%%%%%%%%%%%%%%%%%%%%%%%%%%%%%%%%%%%%%%%%
\subsection{Definition of the quadtree/octree model}
%%%%%%%%%%%%%%%%%%%%%%%%%%%%%%%%%%%%%%%%%%%%%%%%%%%%%%%%%%%%%%%%%%%%%%%%%%%%%%%%
The tree-like structure, called in this paper, quadtree or octree, are a hierarchical data structure based on a recursive subdivision of space \cite{Meagher:1981}. It is particularly useful for representing complex $2-D$ or $3-D$ shapes, and is suitable for Boolean operations like union, difference and intersection. Since the quadtree/octree structure has an implicit adjacency graph, arcwise-connectivity analysis can be naturally achieved. The quadtree/octree model of a space $\cal S$ leads to a representation with cubes of various sizes. Basically, the smallest cubes lie near the boundary of the shape and their size determines the accuracy of the quadtree/octree representation. Quadtree/octrees have been used in several robotic applications \cite{Wenger:1997}, \cite{Faverjon:1984}, \cite{Garcia:1989}.%, \cite{El_Omri:1996}. 

The main advantages of the octree model are (i) very compact file (only B (Black), W (White) and G (Gray) letter) and (ii) accelerates display speed facilities by differentiating model into lower levels of resolution while being rotated and higher resolution while an orientation is temporarily set. 
%In Fig.\ref{figure:quadtree}, we represent an example of quadtree mode with its tree-like representation. 
Conversely, the computational times can be high due to the discretization. If we test only the center of each cube, some singularities can exist and are not detected. This feature is true even if we increase the level of resolution. The accuracy of the model is directly defined by the depth of the quadtree and the size of the initial box. For a quadtree model with a depth $d$ and a initial box of lengths $b$, the accuracy is $b/2^d$.
%\begin{figure}[hbt]
%    \begin{center}
% 				\epsfig{file = quadtree.eps, height= 10mm}
% 				\epsfig{file = exemple_quadtree.eps, height= 10mm}
%        \caption{Quadtree model with white, gray and black nodes on the left and its representation on the %right}
%        \protect\label{figure:quadtree}
%    \end{center}
%\end{figure}
%%%%%%%%%%%%%%%%%%%%%%%%%%%%%%%%%%%%%%%%%%%%%%%
\subsection{Notion of aspect for fully parallel manipulators}
%%%%%%%%%%%%%%%%%%%%%%%%%%%%%%%%%%%%%%%%%%%%%%%
We recall here briefly the definition of the generalized octree defined in \cite{Chablat:1998}:
\begin{Def}
\label{definition:Aspect}
The generalized aspects $\negr A_{ij}$ are defined as the maximal sets in $W \cdot Q$ so that
$\negr A_{ij} \subset W \cdot Q$, $\negr A_{ij}$ is connected, and $\negr A_{ij} = \left\{
                   (\negr X, \negr q) \in Mf_i \setminus det(\negr A) \neq 0
                  \right\}$
\end{Def}
In other words, the generalized aspects $\negr A_{ij}$ are the maximal singularity-free domains of the Cartesian product of the reachable workspace with the reachable joint space.
\begin{Def}
The projection of the generalized aspects in the workspace yields the parallel aspects ${\bf WA}_{ij}$ so that
${\bf WA}_{ij} \subset W$, and ${\bf WA}_{ij}$ is connected.
\end{Def}
The parallel aspects are the maximal singularity-free domains in the workspace for one given working mode.
\begin{Def}
The projection of the  generalized aspects in the joint space yields the serial aspects ${\bf QA}_{ij}$ so that
${\bf QA}_{ij} \subset Q$, and ${\bf QA}_{ij}$ is connected.
\end{Def}
The serial aspects are the maximal singularity-free domains in the joint space for one given working mode.

The aim of this paper is to compute separately the parallel and serial aspects thanks to the properties of the quadtree models.
%%%%%%%%%%%%%%%%%%%%%%%%%%%%%%
\subsection{Introduction to ALIAS library}
%%%%%%%%%%%%%%%%%%%%%%%%%%%%%%
An algorithm for the definition of the joint space and workspace of parallel mechanism is described in the following sections. This algorithm uses the {\tt ALIAS} library~\cite{Merlet:2000}, which is a C++ library of algorithms based on interval analysis. These algorithms deal with systems of
equations and inequalities of which expressions are an arbitrary combination of the most classical mathematical functions (algebraic terms, sine, cosine, log etc..) and of which coefficients are real numbers or, in some cases, intervals. 
Unfortunately, this library is not connected to the octree model and generates large data file to describe by a set of boxes the solution of the problem. Thus, the operations between the set of boxes is more difficult if we compare to the boolean operations that we can made with the octree models.
%%%%%%%%%%%%%%%%%%%%%%%%%%%%%%%%%%%%%%%%%%%%%%%%%%%%%%%%%%%%%%%%%%%%%%%%%%%%%%%%
\subsection{A first basic tool: Box verification}
%%%%%%%%%%%%%%%%%%%%%%%%%%%%%%%%%%%%%%%%%%%%%%%%%%%%%%%%%%%%%%%%%%%%%%%%%%%%%%%%
Our purpose is to determine the quadtree model associated with the joint space or the workspace, that we will call only the space $\cal C$. 
For a given box $B$ defined by two intervals, we note {\it valid box} if it is included in $\cal C$ and {\it invalid
box} otherwise. For that purpose we need to design first a procedure, called ${\cal M}(B)$, that takes as input a box $B$ and returns:
\begin{itemize}
 \item 1: if every point in $B$ is valid,
 \item -1: if no point in $B$ is valid,
 \item 0: if neither of the other two conditions could be verified.
\end{itemize}
To check if one point is valid, we can used several approaches. For example, in the Cartesian space, we have to compute the inverse kinematic model to test if at least one solution exists, which defines the workspace. Thus, for each solution, we can define completely the mechanism for each working mode and compute the determinant of the parallel Jacobian matrix \cite{Chablat:1998}. This procedure is able to define the non-singular domains but their are not necessary connected. However, with a quadtree model, it is easy to perform a connectivity analysis to separate the quadtree model in connected domains. This is a main advantage of the tree-like structures in comparison with the method implemented in the {\tt ALIAS} library. Some additional constraints can be to define dexterous domains in using a kinetostatic index as in \cite{Chablat:2004}.

The problem is now to implement an inverse and direct kinematic model able to take as input a box $B$ and to return, if the box is valid, a box $S$ which contains the solutions of the problem.
%%%%%%%%%%%%%%%%%%%%%%%%%%%%%%%%%%%%%%%%%%%%%%%%%%%%%%%%%%%%%%%%%%%%%%%%%%%%%%%%
\subsection{A second basic tool: Quadtree model definition}
%%%%%%%%%%%%%%%%%%%%%%%%%%%%%%%%%%%%%%%%%%%%%%%%%%%%%%%%%%%%%%%%%%%%%%%%%%%%%%%%
The definition of a quadtree is made recursively by calling several times the same procedure, call ${\cal Q}(B, d, P)$, that takes as input a box $B$, the local depth $d$ in the tree and a pointer $P$ on the data structure which contains the quadtree model.
\begin{itemize}
\item If ${\cal M}(B)$ returns $-1$, it is the end of the recursive search.
\item If ${\cal M}(B)$ returns $1$, a black node is created in the current depth.
\item If ${\cal M}(B)$ returns $0$, if the local depth is smallest than the maximal depth of the quadtree, we divide the box $B$ into four new boxes $B_1$,  $B_2$, $B_3$ and $B_4$, and we call
\end{itemize}
\begin{equation*}
{\cal Q}(B_1, d+1, P) \quad {\cal Q}(B_2, d+1, P) \quad {\cal Q}(B_3, d+1, P) \quad {\cal Q}(B_4, d+1, P) 
\end{equation*}
When ${\cal M}(B)$ returns $-1$, we can also create a another quadtree model, called {\it complementary space} in which we save the box $B$ where at least one constraint is not valid. With this knowledge, we can compute the quadtree model with a higher definition without retesting the valid and non valid boxes. An example of this result will be given in the following section.
%%%%%%%%%%%%%%%%%%%%%%%%%%%%%%%%%%%%%%%%%%%%%%%%%%%%%%%%%%%%%%%%%%%%%%%%%%%%%%%%
\section{Mechanism under study}\label{example}
\begin{figure}%[hbt]
  \begin{center}
    \begin{tabular}{cc}
       \begin{minipage}[t]{52 mm}
       \begin{center}
  				\epsfig{file = 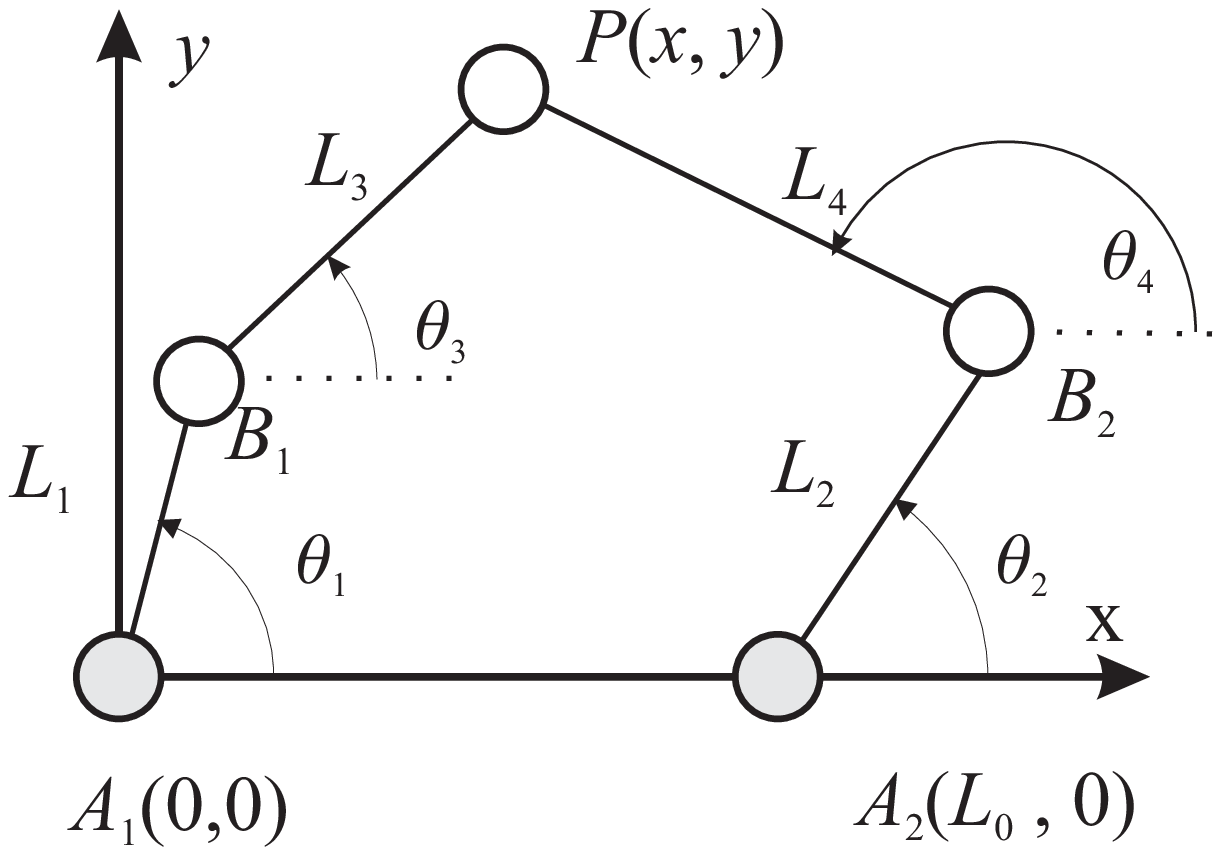, scale= 0.3}
          \caption{A two-dof closed-chain manipulator}
          \protect\label{figure:manipulateur_general}
       \end{center}
       \end{minipage}
       \begin{minipage}[t]{52 mm}
       \begin{center}
    		 \epsfig{file = 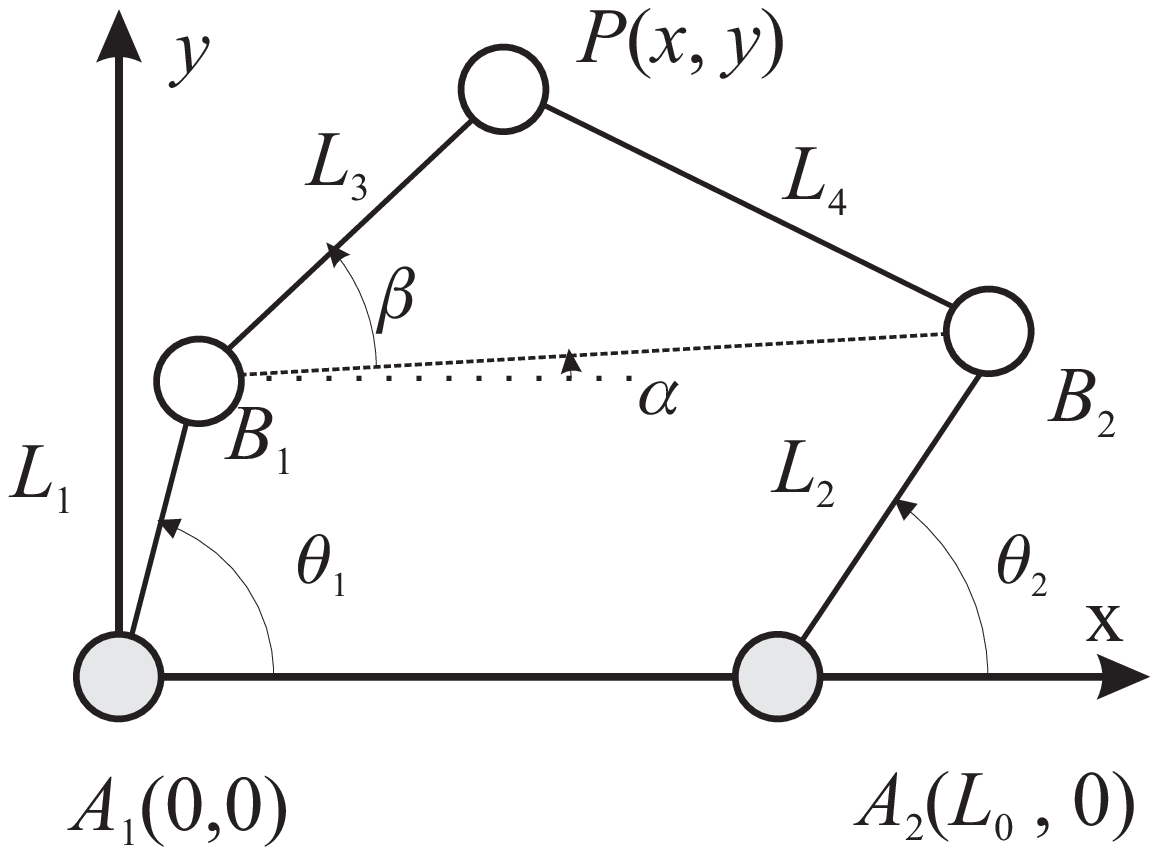, scale= 0.3}
         \caption{The angle $\alpha$ and  $\beta$ used to solve the DKP}
         \protect\label{figure:MGD}
       \end{center}
       \end{minipage}
    \end{tabular}
  \end{center}
\end{figure}

%%%%%%%%%%%%%%%%%%%%%%%%%%%%%%%%%%%%%%%%%%%%%%%%%%%%%%%%%%%%%%%%%%%%%%%%%%%%%%%%
For more legibility, a planar manipulator is used as illustrative example in this paper. This is a five-bar, revolute ($R$)-closed-loop linkage, as displayed in figure~\ref{figure:manipulateur_general}. The actuated joint variables are $\theta_1$ and $\theta_2$, while the Output values are the ($x$, $y$) coordinates of the revolute center $P$. The displacement of the passive joints will always be assumed unlimited in this study. Lengths $L_0$, $L_1$, $L_2$, $L_3$, and $L_4$ define the geometry of this manipulator entirely.

Two sets of dimensions are used to illustrate the algorithm. The first one, called ${\cal M}_1$, is defined in \cite{Chablat:1998} and its dimensions are  $L_0~=~9$, $L_1~=~8$, $L_2~=~5$, $L_3~=~5$ and $L_4~=8~$, in certain units of length that we need not to specify. And, the second one, called ${\cal M}_2$, is defined in \cite{Bonev:2007} and its dimensions are $L_0~=~2.55$, $L_1~=~2.3$, $L_2~=~2.3$, $L_3~=~2.3$ and $L_4~=2.3~$.

\subsection{Kinematic Relations}
The velocity $\dot{\bf p}$ of point $P$, of position vector {\bf p}, can be obtained in two different forms, depending on the direction in
which the loop is traversed, namely,
%\begin{subequations}
\begin{equation} %Equation cinematique du modele du manipulateur RR-RRR
        \dot{\bf p}= \dot{\bf b_1} + \dot{\theta}_3 {\bf E}
                     ({\bf p}  - {\bf b_1}) \quad
%        \protect\label{equation:kinematic_1}
        \dot{\bf p}= \dot{\bf b_2} + \dot{\theta}_4 {\bf E}
                     ({\bf p}  - {\bf b_2})
        \protect\label{equation:kinematic_2}
\end{equation}
with matrix ${\bf E}$ defined as $
{\bf E}= \left[\begin{array}{cc}
              0 & -1 \\
              1 &  0
             \end{array}
        \right]$
and ${\bf b}_1$ and ${\bf b}_2$ denoting the position vectors, in the frame indicated in figure~\ref{figure:manipulateur_general} of points $B_1$ and $B_2$, respectively. Furthermore, note that $\dot{\bf c}$ and $\dot{\bf d}$ are given by
\begin{eqnarray}
\dot{\bf b_1}= \dot{\theta}_1 {\bf E}({\bf b}_1 -  {\bf a}_2), \quad
\dot{\bf b_2}= \dot{\theta}_2 {\bf E}({\bf b}_2 -  {\bf a}_2)   \nonumber
\end{eqnarray}
Two Jacobian matrix, ${\bf A}$  and {\bf B}, permit to study the serial and parallel singular configurations \cite{Wenger:1997},
\begin{equation}
{\bf A}= \left[\begin{array}{c}
                ({\bf p}  - {\bf b_1})^T \\
                ({\bf p}  - {\bf b_2})^T
              \end{array}
         \right] 
         {,~}
         {\bf B}\!\!=  \!\!
					\left[\begin{array}{cc}
                   \!\!L_1 L_2 \sin(\theta_3 - \theta_1) \!\!\!\!\!\!\!&
                   0                                 \\
                   0                                 &
                   \!\!\!\!\!\!\!L_3 L_4 \sin(\theta_4 - \theta_2)\!\!
                \end{array}
           \right] \!\!
           = \!\!
           \left[\begin{array}{cc}
                   \!B_{11} \!\!&
                   0                                 \\
                   0                                 &
                   \!\!B_{22}\!
                \end{array}
           \right]
       \protect\label{equation:jacobian_matrices_B}
\end{equation}
Two assembly modes can be defined with the sign of $\det(\bf A)$. To characterize the assembly mode, we can only compute $({\bf p}  - {\bf b_1}) \times ({\bf p}  - {\bf b_2})$.
%\begin{figure}%[hbt]
%    \begin{center}
% 				\epsfig{file = mode_assemblage.eps, scale= 0.25}%
%				\vspace {-4mm}
%        \caption{The two assembly modes and the singular configurations}
%        \protect\label{figure:assembly_modes}
%        \vspace {-4mm}
%    \end{center}
%\end{figure}

The parallel singular configurations are located at the boundary of the joint space. Such singularities occur whenever $B_1$, $P$ and $B_2$ are aligned. In such configurations, the manipulator cannot resist an effort in the orthogonal direction of $B_1B_2$. Besides, when $B_1$ and $B_2$ coincide, the position of $P$ is no longer controllable since $P$ can rotate freely around $B_1$ even if the actuated joints are locked. This singularity cannot be find by the discretization of the workspace but is detected by the interval analysis based method.
%The four working modes depicted in Fig.~\ref{figure:working_modes} can be defined by the sign of $B_{11}$ and $B_{22}$. 
%\begin{figure}[hbt]
%    \begin{center}
% 				\epsfig{file = mode_fonctionnement.eps, scale= 0.25}
%				\vspace {-4mm}
%        \caption{The four working modes}
%        \protect\label{figure:working_modes}
%    \end{center}
%\end{figure}
%%%%%%%%%%%%%%%%%%%%%%%%%%%%%%%%%%%%%%%%%%%%%%%%%%%%%%%%%%%%%%%%%%%%%%%%%%
\subsection{Direct kinematic problem}
%%%%%%%%%%%%%%%%%%%%%%%%%%%%%%%%%%%%%%%%%%%%%%%%%%%%%%%%%%%%%%%%%%%%%%%%%%
For planar mechanism, the computation of the direct kinematic problem (DKP) is very simple. However, if we want to return the good information to the function $\cal C$, we have to distinguish more cases. If there is no joint limits, they are zero or two solutions for the DKP.

The procedure takes as input a box $B$ defined by two intervals $\tilde{\theta_1}=[\underline{\theta_1}, \overline{\theta_1}]$ and $\tilde{\theta_2}= [\underline{\theta_2}, \overline{\theta_2}]$, the actuated joint variables. In this example, the values of the length are defined as a float but can be also defined by a interval to represent the tolerances of manufacturing.

The algorithm can be described by the following steps: 
\begin{itemize}
\item Compute the position of $\tilde{\negr b_1}$ and $\tilde{\negr b_2}$ which is the location of $B_1$ and $B_2$ respectively.
\item Compute the distance $\tilde{L}$ between $B_1$ and $B_2$:
\begin{itemize}
  \item $[\underline{L}, \overline{L}]= || \tilde{\negr b_1} - \tilde{\negr b_2}||$
  \item  If $\underline{L} > L_3+L_4$ then the mechanism can be assembly and the function return -1. \par
  \item  If $\underline{L} = 0$ then the mechanism can be in a singular configuration. 
\end{itemize}
\item Compute the angle $\alpha$ and $\beta$ of the triangle ($B_1$, $B_2$, $P$) (Figure~\ref{figure:MGD}).
\begin{itemize}
\item $C(\alpha)= \sqrt{(\tilde{L}^2 + L_3^2 - L_4^2) / (2  L_3  \tilde{L})}. $
\item $\beta= \arctan \left(\frac{B_{2y} - B_{1y}} {B_{2x} - B_{1x}})\right)$
\item If ($\underline{C(\alpha)} \geq -1$ and $C(\alpha) \leq 1$) then 
$\alpha= \arccos(C(\alpha))$.
\end{itemize}
\item Compute the two solutions $P_1$ and $P_2$ of point $P$:
\begin{itemize}
\item $P_{1x}= B_{1x} + L_3 \cos(\alpha+\beta)$
\item $P_{1y}= B_{1y} + L_3 \sin(\alpha+\beta)$
\item $P_{2x}= B_{1x} + L_3 \cos(\alpha-\beta)$
\item $P_{2y}= B_{1y} + L_3 \sin(\alpha-\beta)$.
\end{itemize}
\end{itemize}
To perform this procedure, all the trigonometric function come from the {\tt ALIAS} library and accept as input an interval and return an interval. For each step, we have to check that the argument are not out of range. 

\begin{figure}%[hbt]
    \begin{center}
    \begin{tabular}{cc}
       \begin{minipage}[t]{52 mm}
       \begin{center}
 				\epsfig{file = 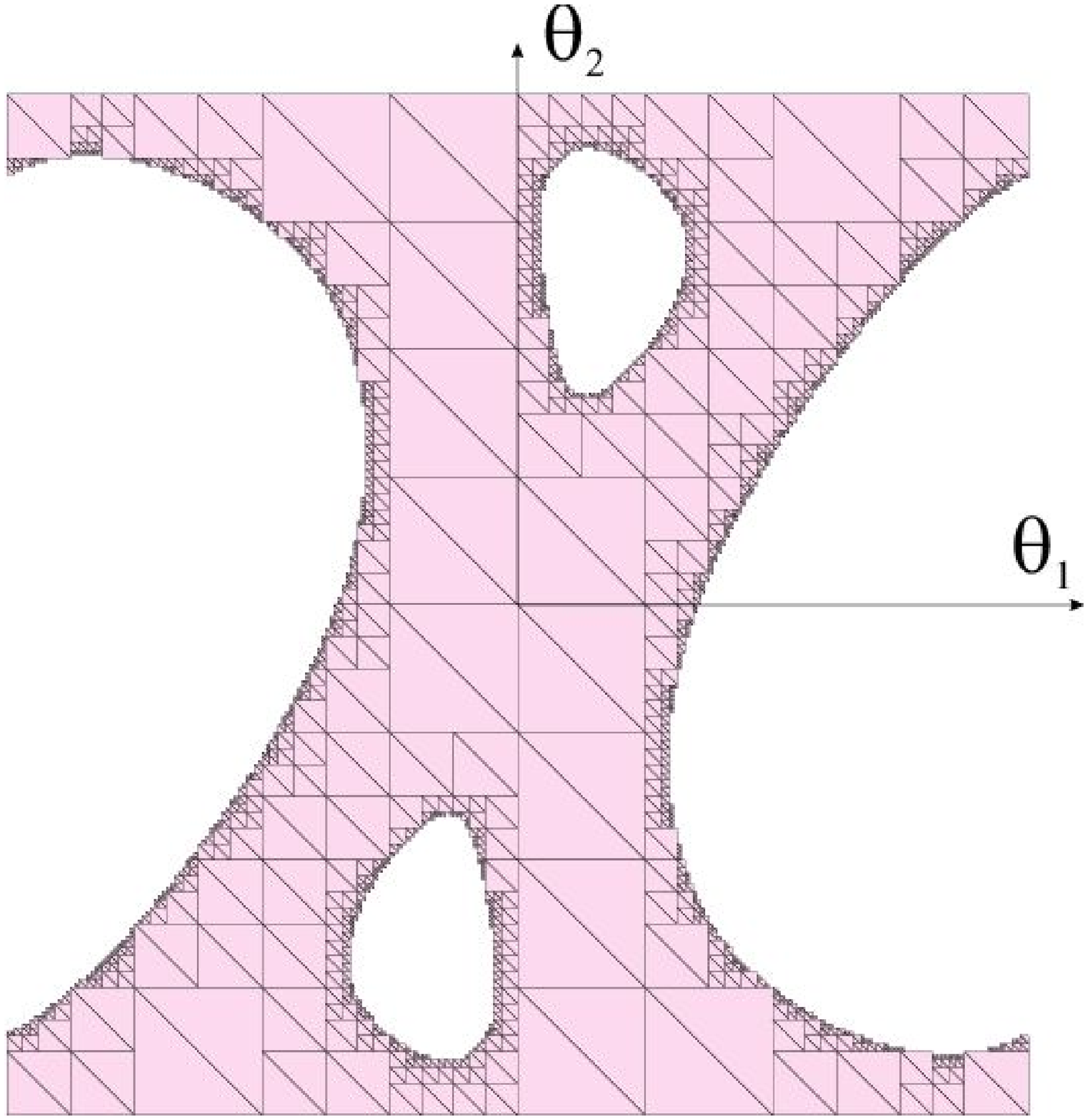, scale= 0.13}
 				\epsfig{file = 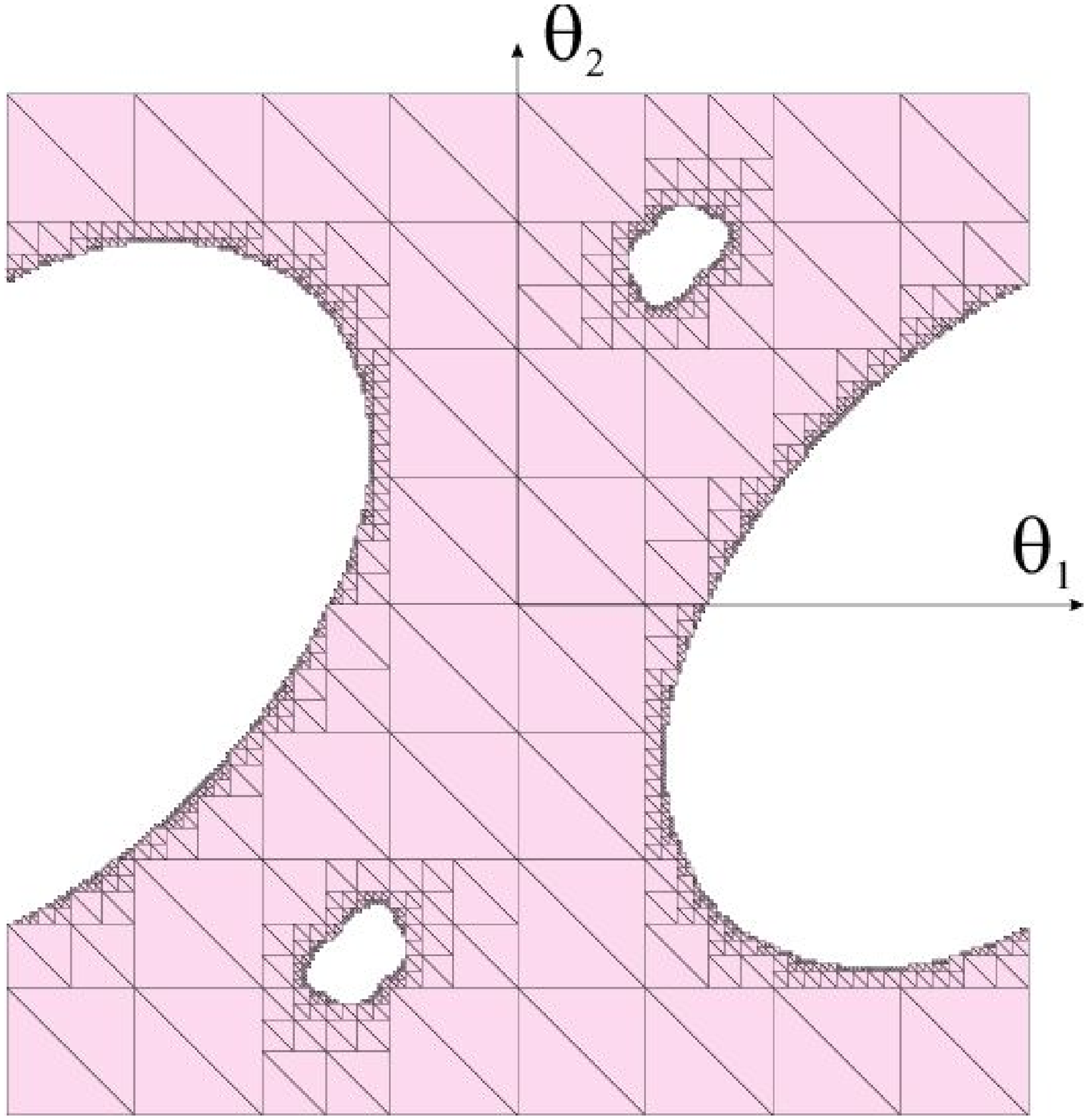, scale= 0.13}
        \caption{Joint space of ${\cal M}_1$ and ${\cal M}_2$}
        \protect\label{figure:joint_space_damien}
       \end{center}
     \end{minipage} &
     \begin{minipage}[t]{52 mm}
       \begin{center}
 				\epsfig{file = 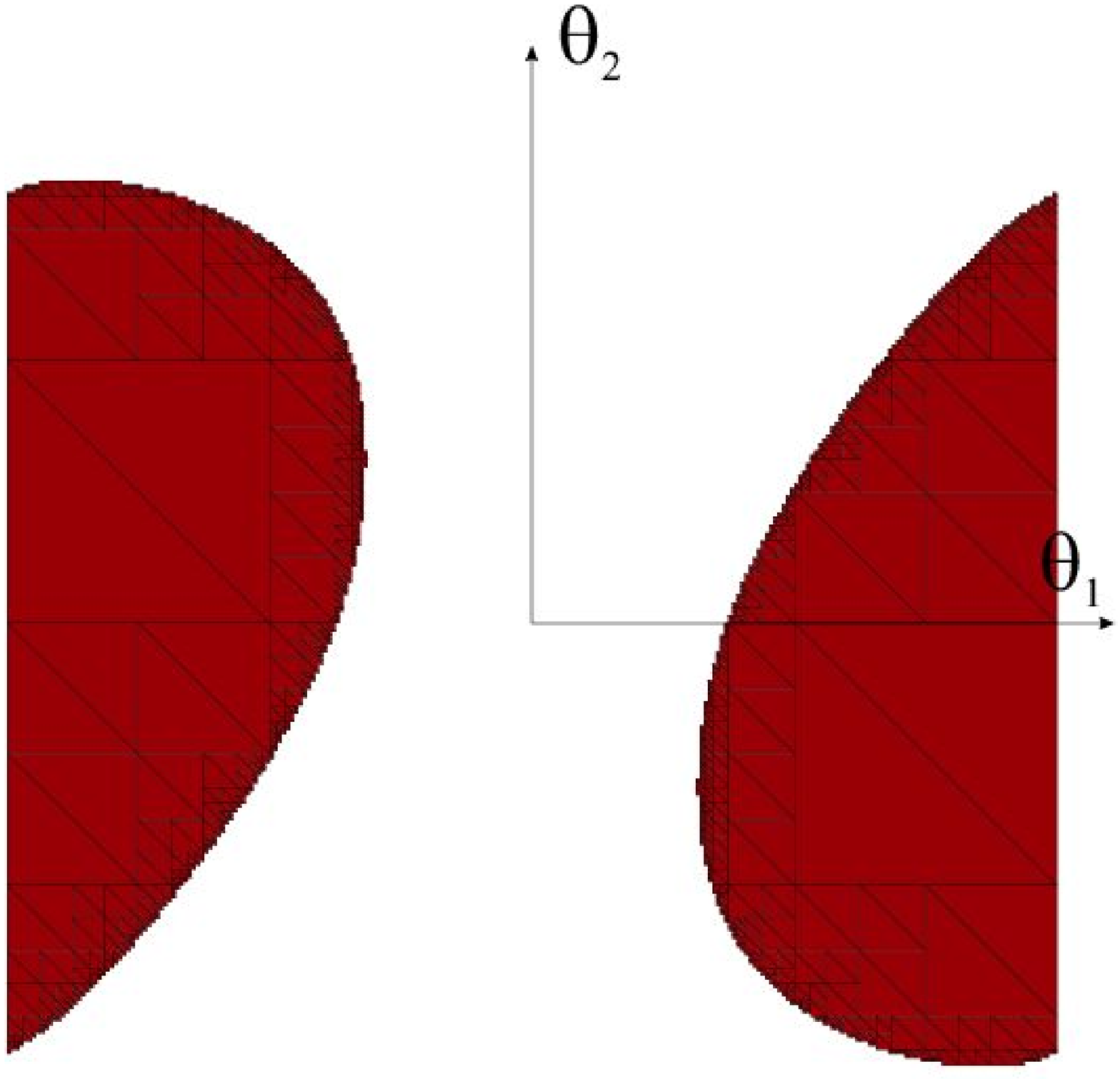, scale= 0.13}
 				\epsfig{file = 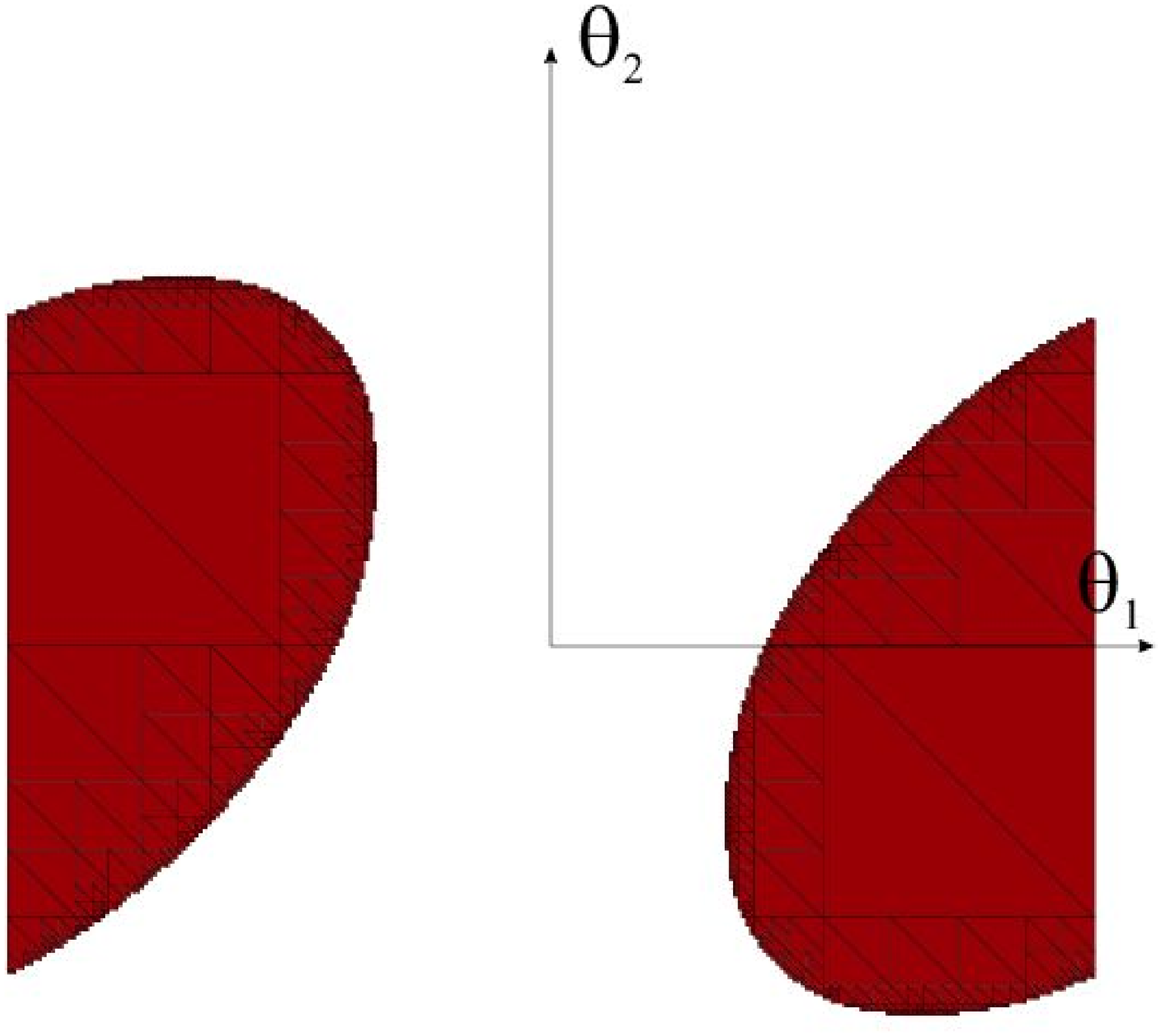, scale= 0.13}
        \caption{The complementary joint space of ${\cal M}_1$ and ${\cal M}_2$}
        \protect\label{figure:joint_space_damien_vide}
       \end{center}
     \end{minipage}
     \end{tabular}
    \end{center}
\end{figure}

For the two mechanism under study, we can plot with this function the joint space (Figure~\ref{figure:joint_space_damien}). The results are equivalent to the solution computed in \cite{Chablat:1998} and \cite{Bonev:2007}. However, conversely to a discretization method, we can detect the parallel singularity where $B_1$ and $B_2$ coincide. Normally, it is a point but with the interval analysis method, the algorithm cannot define valid boxec where $\det(\negr A)$ is equal to zero.	
The set of box where at least one constraint is not valid  is called the complementary joint space, figure~\ref{figure:joint_space_damien_vide}.  
%%%%%%%%%%%%%%%%%%%%%%%%%%%%%%%%%%%%%%%%%%%%%%%%%%%%%%%%%%%%%%%%%%%%%%%%%%%%
\subsection{Selection of the assembly mode}
%%%%%%%%%%%%%%%%%%%%%%%%%%%%%%%%%%%%%%%%%%%%%%%%%%%%%%%%%%%%%%%%%%%%%%%%%%%%
For a parallel mechanism with only two solutions to the direct kinematic problem, the assembly mode is characterize by the sign of $\det(\negr A)$. To simplify, we can only compute
\be
    \tilde{\negr t}= (\tilde{\negr b_1} - \tilde{\negr p}) \times (\tilde{\negr b_2} - \tilde{\negr p})
\ee
and to test if $\tilde{\negr t_z}$ is positive or negative.
%%%%%%%%%%%%%%%%%%%%%%%%%%%%%%%%%%%%%%%%%%%%%%%%%%%%%%%%%%%%%%%%%%%%%%%%%%%%
\subsection{Inverse kinematic problem}
%%%%%%%%%%%%%%%%%%%%%%%%%%%%%%%%%%%%%%%%%%%%%%%%%%%%%%%%%%%%%%%%%%%%%%%%%%%%
For the mechanism under study, there are four solutions for the inverse kinematic problem (IKP), two solutions for each leg. %(Figure~\ref{figure:working_modes})
The procedure takes as input a box $B$ defined by to two intervals $\tilde{x}=[\underline{x}, \overline{y}]$ and $\tilde{y}= [\underline{y}, \overline{y}]$, the position of $\tilde{P}$.

\begin{figure}%[hbt]
    \begin{center}
    \begin{tabular}{cc}
       \begin{minipage}[t]{50 mm}
         \begin{center}
        	\epsfig{file = 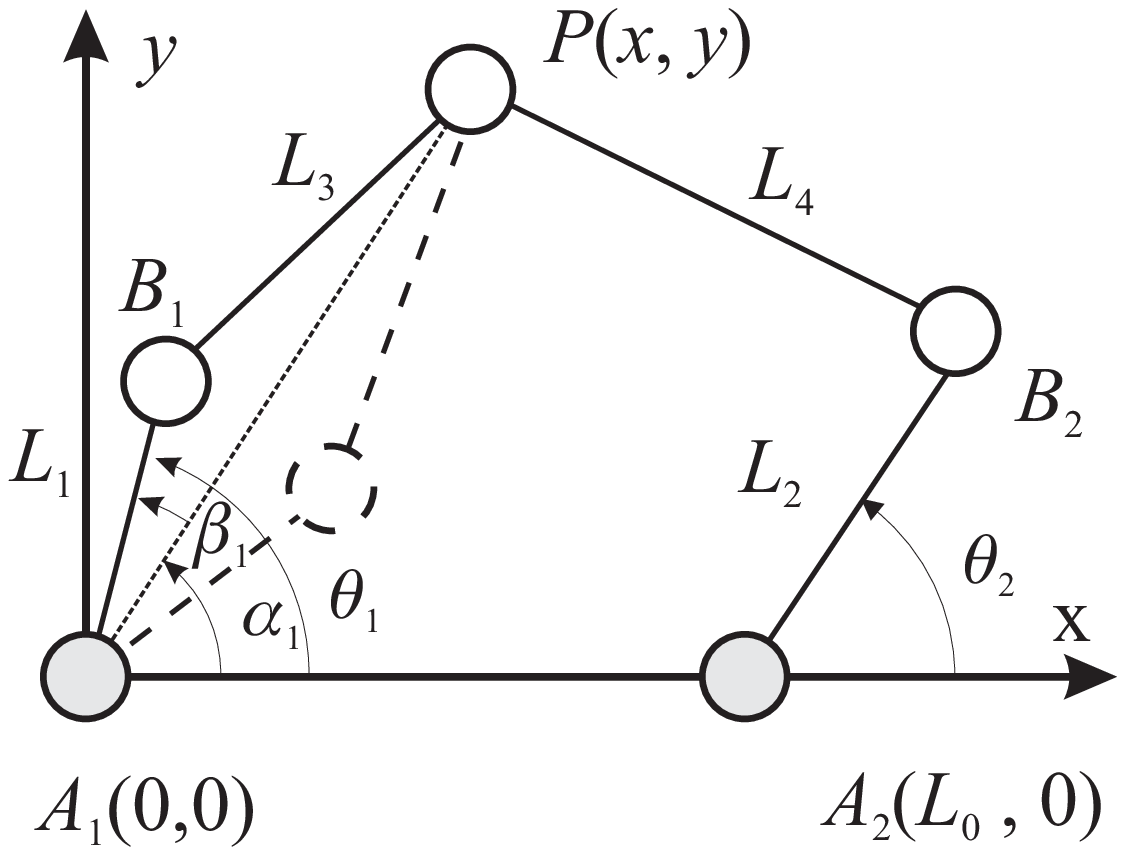, scale= 0.3}
          \caption{The angle $\alpha$ and  $\beta$ used to solve the IKP}
          \protect\label{figure:MGI}
         \end{center}
       \end{minipage} &
       \begin{minipage}[t]{50 mm}
         \begin{center}
 			  	\epsfig{file = 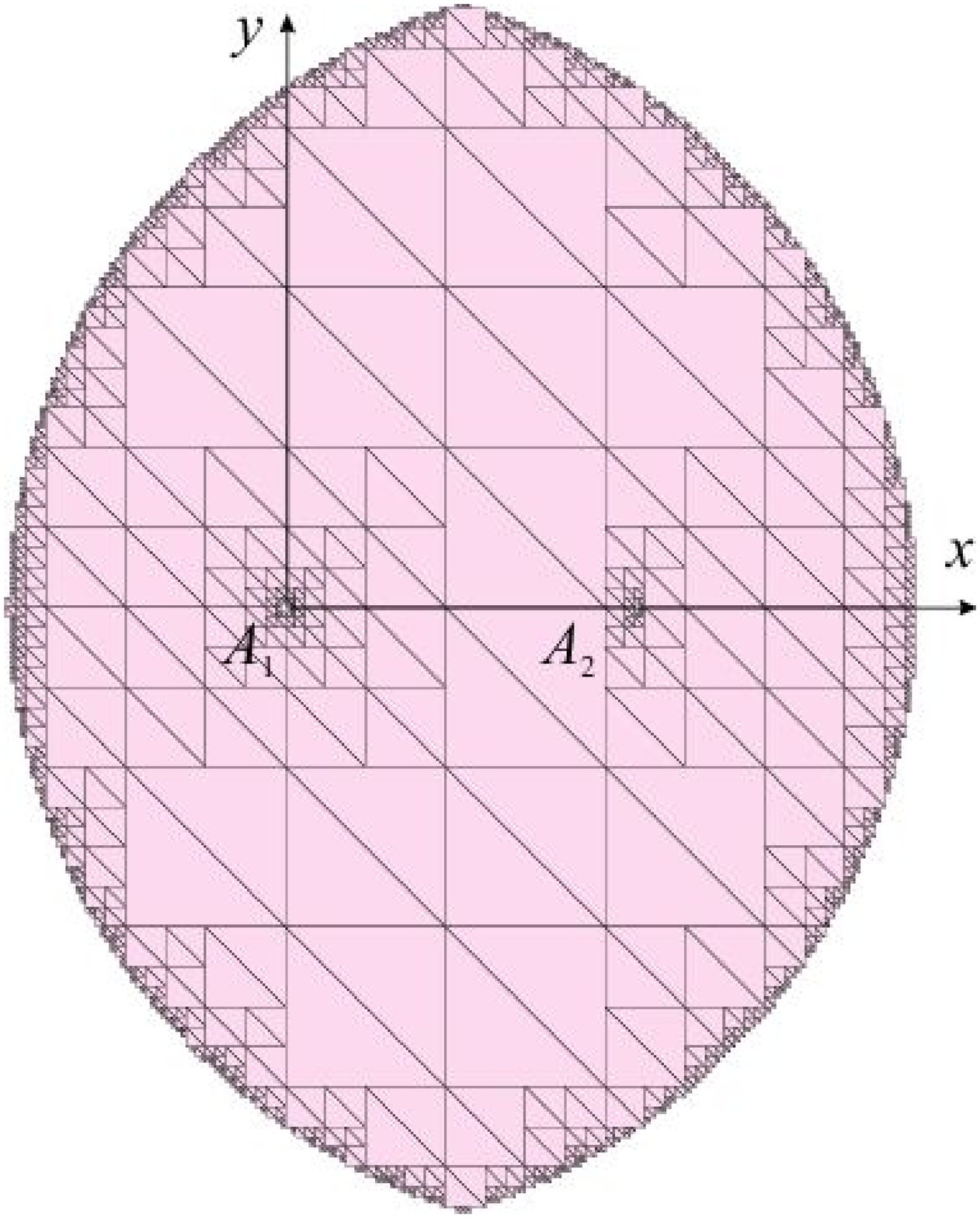,  scale= 0.15}
 		  		\epsfig{file = 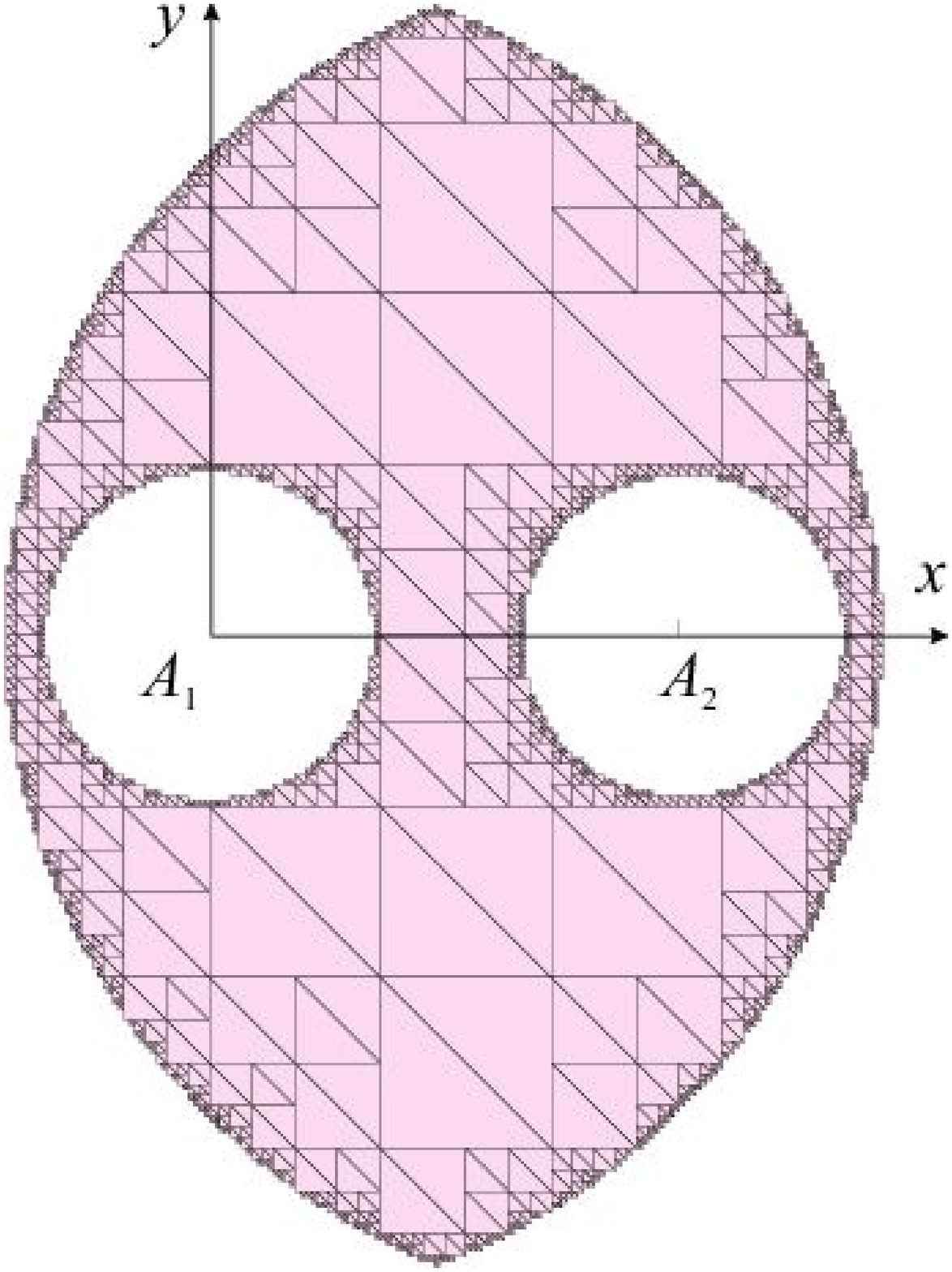, scale= 0.15}
          \caption{Workspace of ${\cal M}_1$ and ${\cal M}_2$}
          \protect\label{figure:workspace_damien}
         \end{center}
       \end{minipage}
    \end{tabular}
    \end{center}
\end{figure}

The algorithm can be described by the following steps: 
\begin{itemize}
\item Compute the distance $\tilde{M}_1$ between $A_1$ and $P$, and $\tilde{M}_2$ between $A_2$ and $\tilde{P}$.
\item If ($\underline{M_1}>L_1+L_3$ or $\underline{M_2}>L_2+L_4$) then $P$ is outside the workspace and the function returns -1.
\item If ($\overline{M_1} < ||L_1-L_3||$ or $\overline{M_2} < ||L_2-L_4||$) then $P$ is located in a hole of the workspace and the function returns -1.  
\item If ($\underline{M_1}> 0$) and ($\underline{M_2}>0$) and ($\underline{M_1}>||L_1-L3||$) and ($\underline{M_2}>||L_2-L4||$) and ($\overline{M_1} < L_1+L_3$) and ($\overline{M_2} < L_2+L_4$) then compute the angles $\beta_1=\widehat{B_1A_1P}$ and $\beta_2=\widehat{B_2A_2P}$ noted in Figure~\ref{figure:MGI}:
\begin{itemize}
\item $\tilde{C}(\beta_1)= \frac{L_1^2 + \tilde{M}_1^2 - L_3^2}{2\tilde{M}_1 L_1}$ \quad $\tilde{C}(\beta_2)= \frac{L_2^2 + \tilde{M}_2^2 - L_4^2}{2\tilde{M}_2 L_2}$
\item If ($||\tilde{C}(\beta_1)|| \leq 1$) and ($||\tilde{C}(\beta_2)|| \leq 1$) then
\begin{itemize}
\item	$\alpha_1= \arctan({P_y/P_x}) \quad \alpha_2= \arctan({P_y/(C_2 - P_x}))$
\item $\beta_1= \arccos(\tilde{C}(\beta_1))$ \quad $\beta_2= \arccos(\tilde{C}(\beta_2))$
\item $\alpha_{11}= \alpha_1 + \beta_1 \quad \alpha_{12}= \alpha_1 - \beta_1 \quad 
      \alpha_{21}= \pi - \alpha_2 + \beta_2 \quad \alpha_{22}= \pi - \alpha_2 - \beta_2$
\item Returns 1;
\end{itemize}
\item Else, returns 0;
\end{itemize}
\end{itemize}

For the two mechanism under study, we can plot with this function the joint space (Figure~\ref{figure:workspace_damien}). For the mechanism $\cal{M}_2$, there is a hole in $A_1$ and $A_2$ which is only a point. On the picture, we can only notice that there exists a sub-division of the space because there are small boxes. The size of the hole is equal to the size of the smallest box.
%%%%%%%%%%%%%%%%%%%%%%%%%%%%%%%%%%%%%%%%%%%%%%%%%%%%%%%%%%%%%%%%%%%%%%%%%%%%
\subsection{Selection of the working mode}
%%%%%%%%%%%%%%%%%%%%%%%%%%%%%%%%%%%%%%%%%%%%%%%%%%%%%%%%%%%%%%%%%%%%%%%%%%%%
For a parallel mechanism with only four solutions to the inverse kinematic problem, the working mode is characterized by the sign of $B_{11}$ and $B_{22}$. This can be done simply 
\be
    \tilde{\negr u}= (\tilde{\negr b_1} - \tilde{\negr a_1}) \times (\tilde{\negr p} - \tilde{\negr b_1}) \quad
    \tilde{\negr v}= (\tilde{\negr b_2} - \tilde{\negr a_2}) \times (\tilde{\negr p} - \tilde{\negr b_2})
\ee
and to test the sign of $\tilde{\negr u_{z}}$ or $\tilde{\negr v_{z}}$.
%%%%%%%%%%%%%%%%%%%%%%%%%%%%%%%%%%%%%%%%%%%%%%%%%%%%%%%%%%%%%%%%%%%%%%%%%%%%
\subsection{Computation of the generalized aspects}
%%%%%%%%%%%%%%%%%%%%%%%%%%%%%%%%%%%%%%%%%%%%%%%%%%%%%%%%%%%%%%%%%%%%%%%%%%%%
The number of aspects is the same for the both mechanisms. We can compute separately the serial and parallel aspects for all the working modes. %Tables~\ref{table:aspect_M1} and \ref{table:aspect_M2} contain the number of aspects associated with their figures. 
For a give sign of $\det(\negr A)$, $B_{11}$ and $B_{22}$, we have define a quadtree model. To obtain the generalized aspects, a connectivity analysis is made to separate the connected regions. We need to test only one point in the workspace and its projection on the joint space to associate a serial aspect to its parallel aspect counterpart. 

\begin{figure}%[hbt]
    \begin{center}
    \begin{tabular}{cc}
       \begin{minipage}[t]{51 mm}
        \center
 				\epsfig{file = 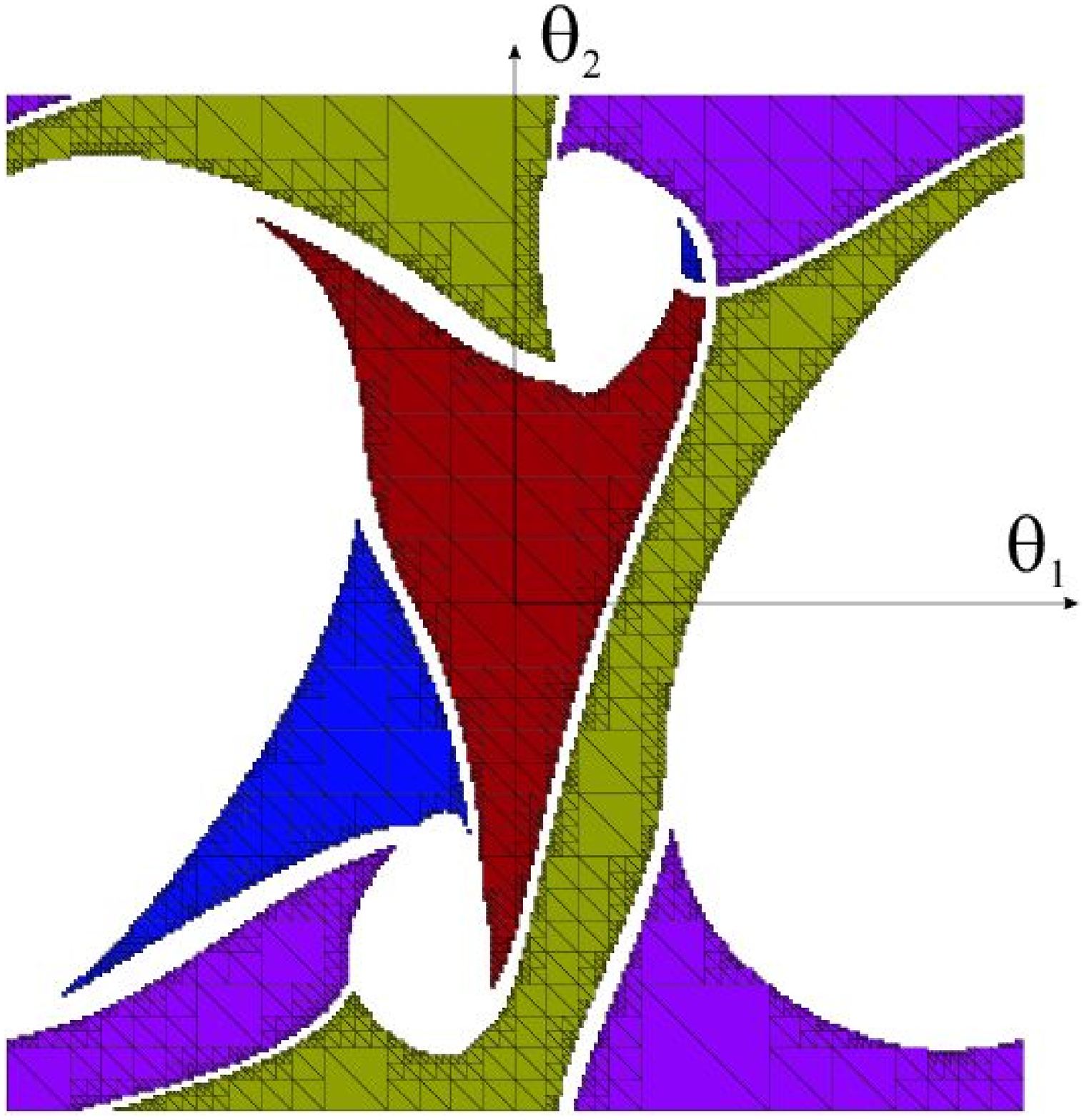, scale= 0.12}
         %\vspace{-4mm}
        \caption{The serial aspects of ${\cal M}_1$ for $\det(\negr A)>0$}
        \protect\label{figure:joint_space_damien_all}
     \end{minipage} &
     \begin{minipage}[t]{51 mm}
        \center
 				\epsfig{file = 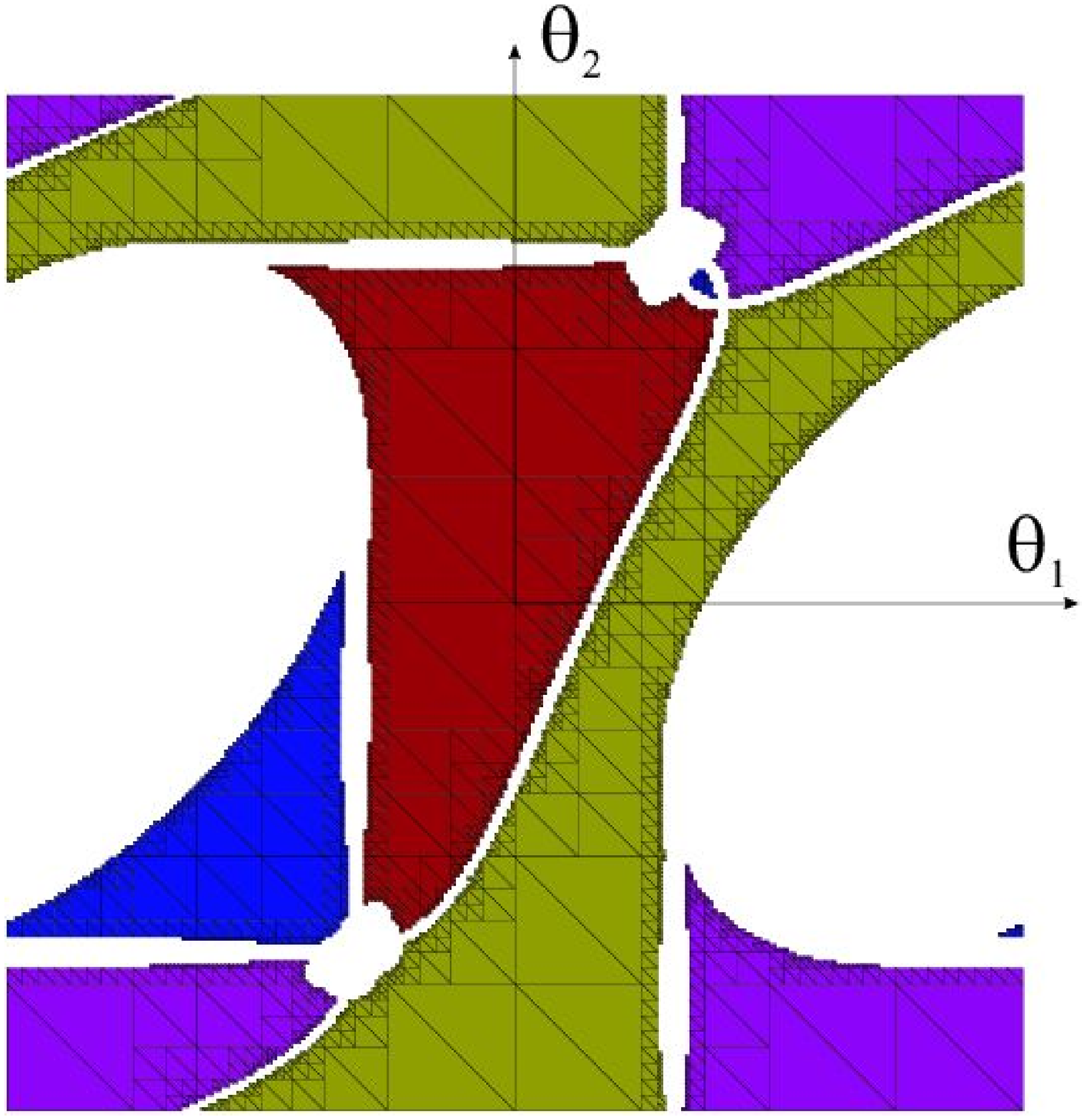, scale= 0.12}
         %\vspace{-4mm}
        \caption{The serial aspects of ${\cal M}_2$ for $\det(\negr A)>0$}
        \protect\label{figure:joint_space_ilian_all}
       \end{minipage}
    \end{tabular}
    \end{center}
\end{figure}

%\begin{table}
%    \begin{center}
%       \caption{The generalized aspects for the mechanism ${\cal M}_1$}
%       \label{table:aspect_M1}
%       \begin{tabular}{|c|c|c|c|c|c|c|c|c|} \hline
%       $Figure~\ref{figure:Aspect_M1}$&(a)&(b)&(c)&(d)&(e)&(f)&(g)& h)\\ \hline
%       $det(\negr A)$ & P & P & P & P & N & N & N & N  \\ \hline
%       $ B_{11}$ & P & P & N & N & P & P & N & N  \\ \hline
%       $ B_{22}$ & P & N & N & P & P & N & N & P  \\ \hline
%       $\begin{array}{c}
%       \!Nb~of~generalized~aspects\!
%       \end{array}$   & 1 & 2 & 1 & 1 & 1 & 1 & 1 & 2 \\ \hline
%       \end{tabular}
%    \end{center}
%\end{table}
%\begin{table}
%    \begin{center}
%       \caption{The generalized aspects for the mechanism ${\cal M}_2$}
%       \label{table:aspect_M2}
%           \begin{tabular}{|c|c|c|c|c|c|c|c|c|} \hline
%       $Figure~\ref{figure:Aspect_M2}$&(a)&(b)&(c)&(d)&(e)&(f)&(g)& h)\\ \hline
%       $det(\negr A)$ & P & P & P & P & N & N & N & N  \\ \hline
%       $ B_{11}$ & P & P & N & N & P & P & N & N  \\ \hline
%       $ B_{22}$ & P & N & N & P & P & N & N & P  \\ \hline
%       $\begin{array}{c}
%       \!Nb~of~generalized~aspects\!
%       \end{array}$   & 1 & 2 & 1 & 1 & 1 & 1 & 1 & 2 \\ \hline
%       \end{tabular}
%    \end{center}
%\end{table}
The study of the joint space allow us to know if a trajectory between two generalized aspects exists by passing through a serial singularity. Figures~\ref{figure:joint_space_damien_all} and \ref{figure:joint_space_ilian_all} permit us to conclude that, for ${\cal M}_1$ and ${\cal M}_2$, no trajectory exists in which we change only one time the working modes, between the aspects 1 and 4 and between the aspects 2, 3 and 5.
\begin{figure}
    \begin{center}
 				\epsfig{file = 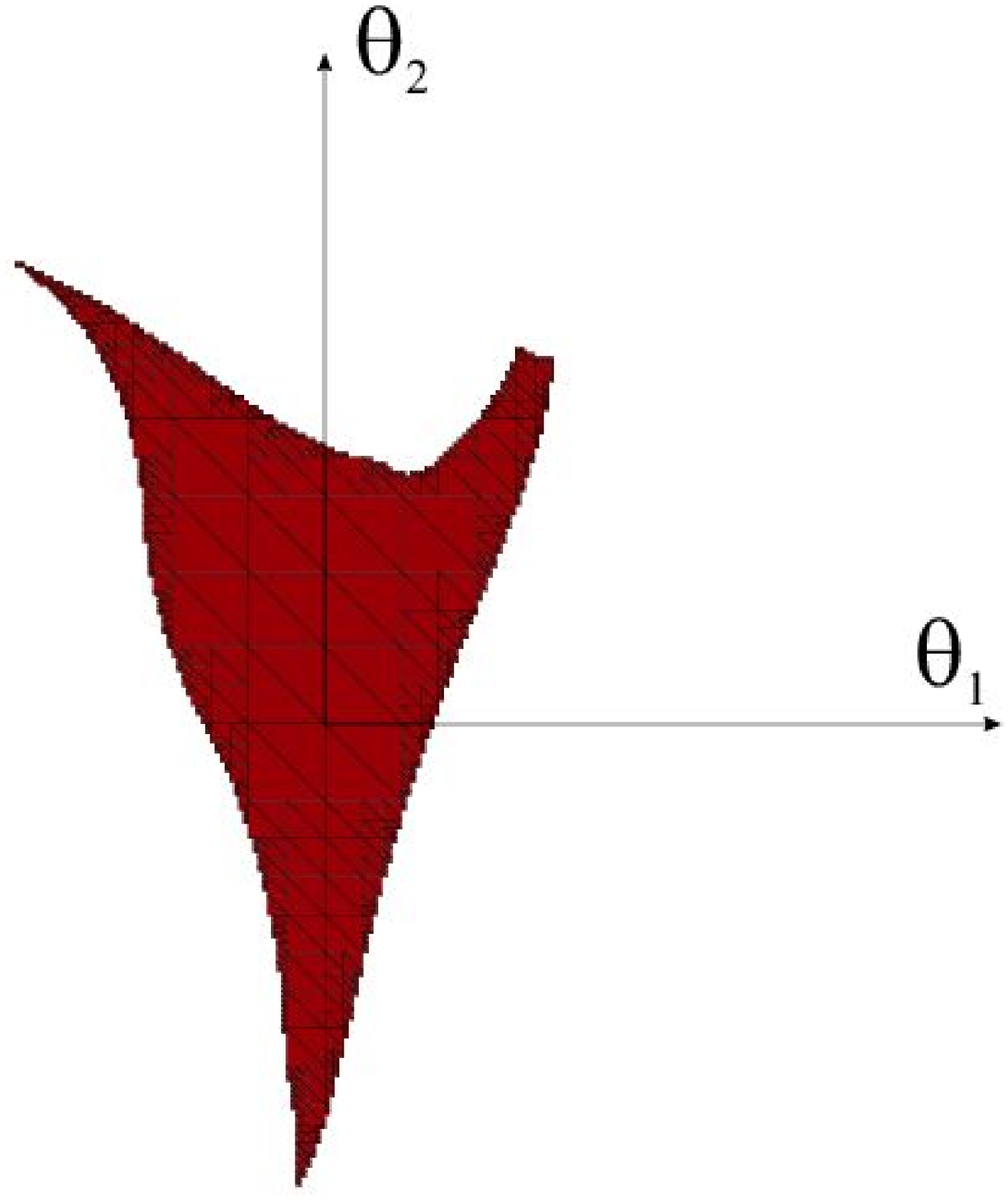, scale= 0.07}
 				\epsfig{file = 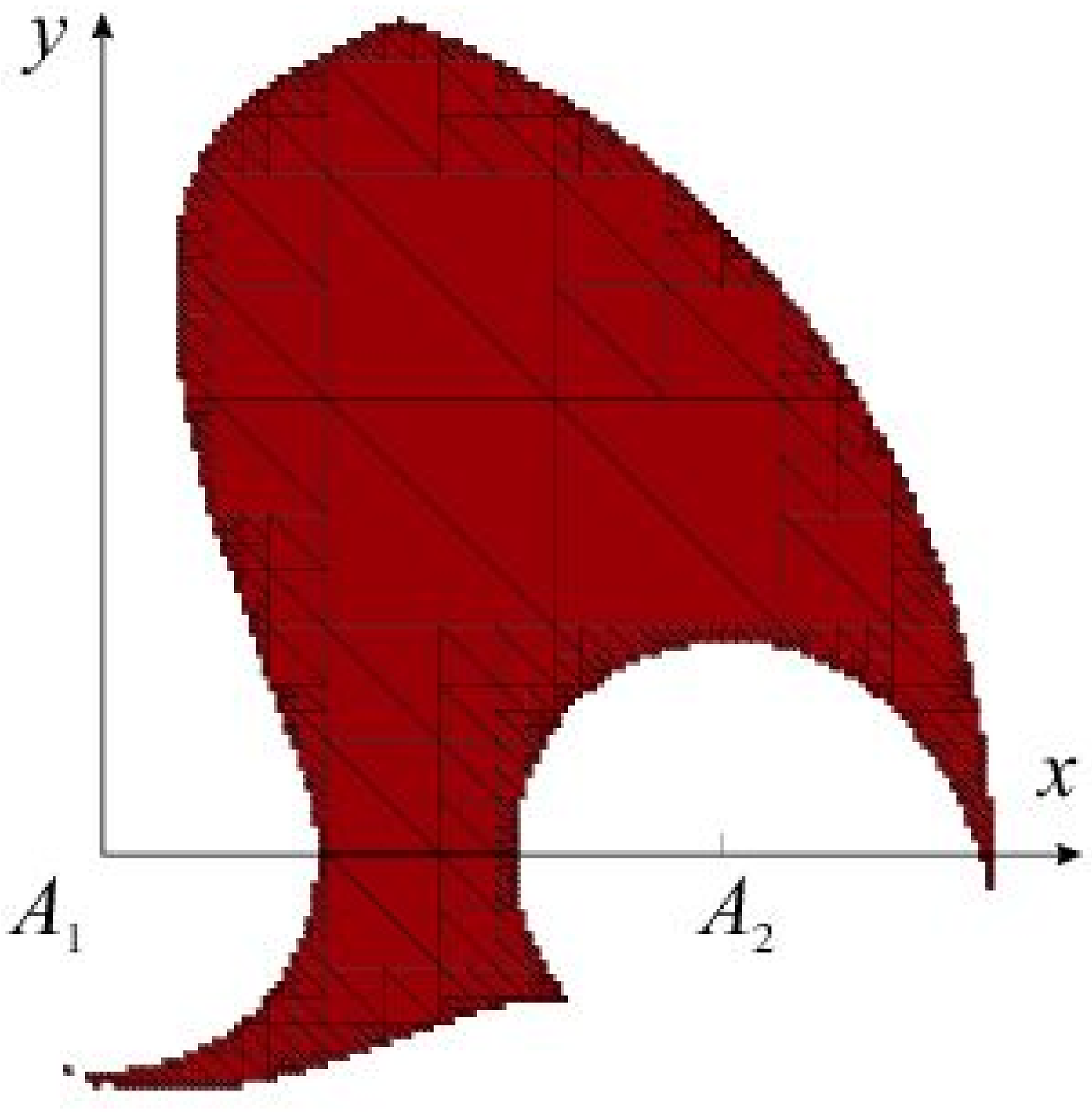, scale= 0.07}\small{(a)}
 				\epsfig{file = 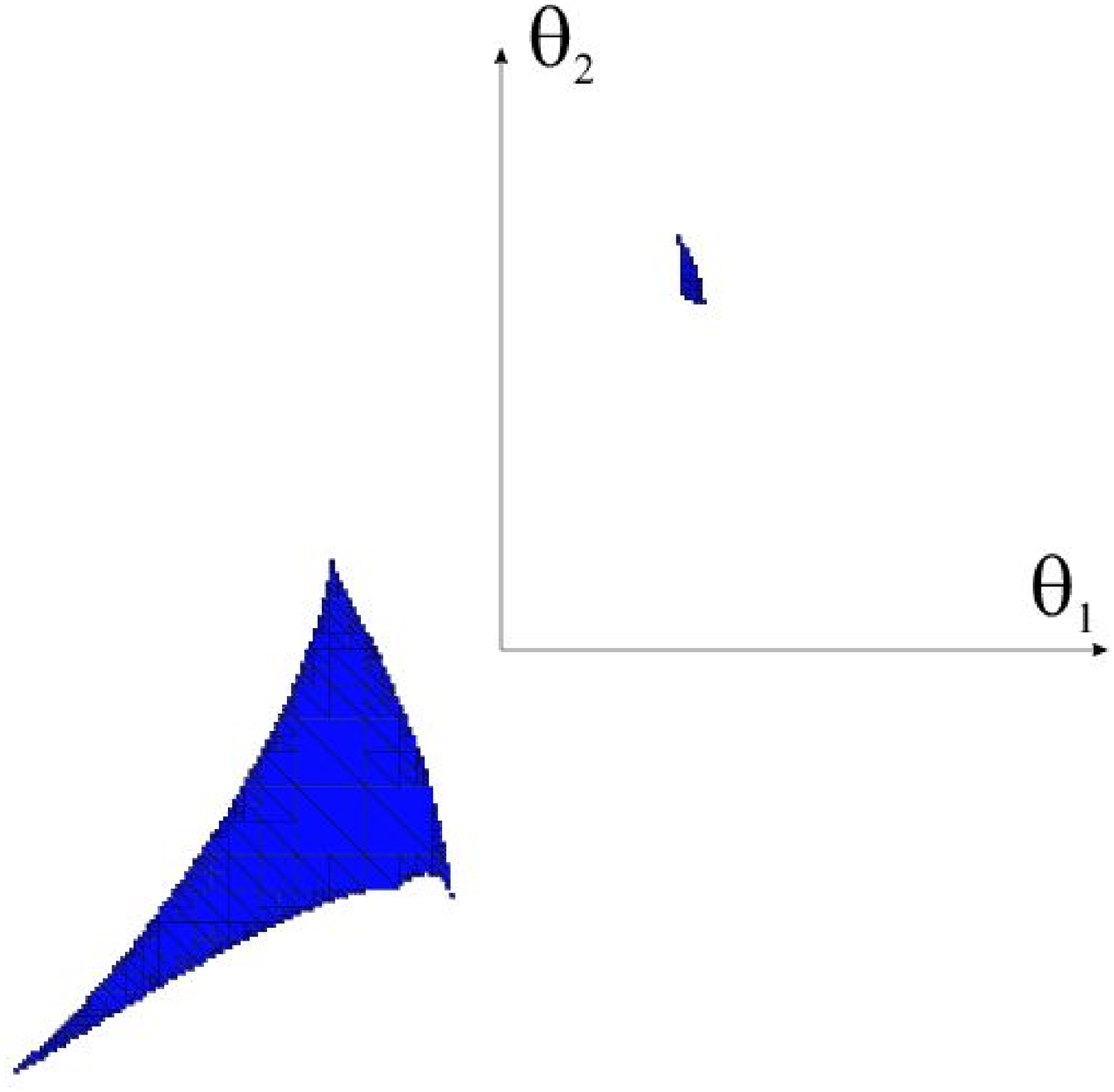, scale= 0.07}
 				\epsfig{file = 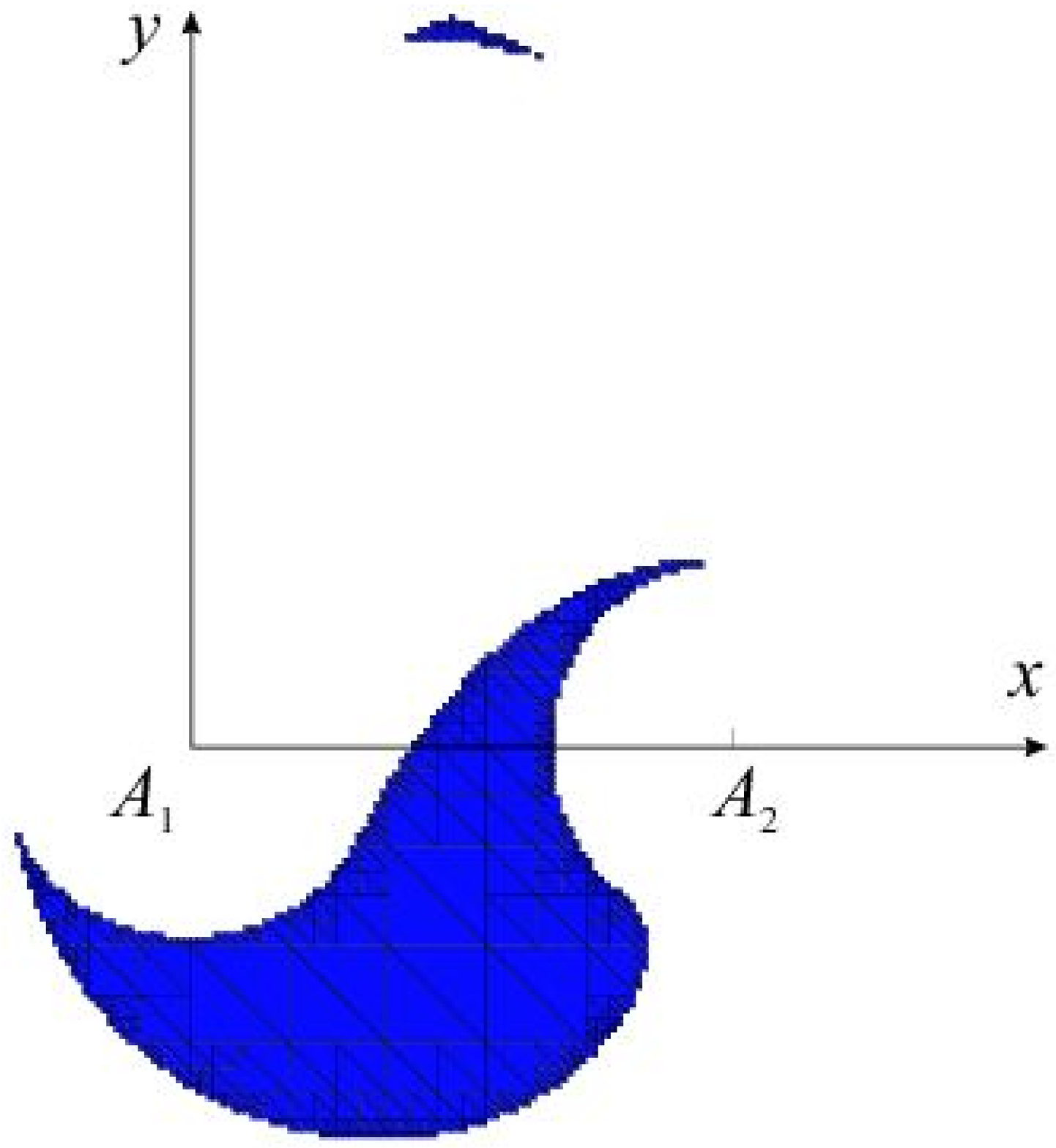, scale= 0.07}\small{(b)}
 				\epsfig{file = 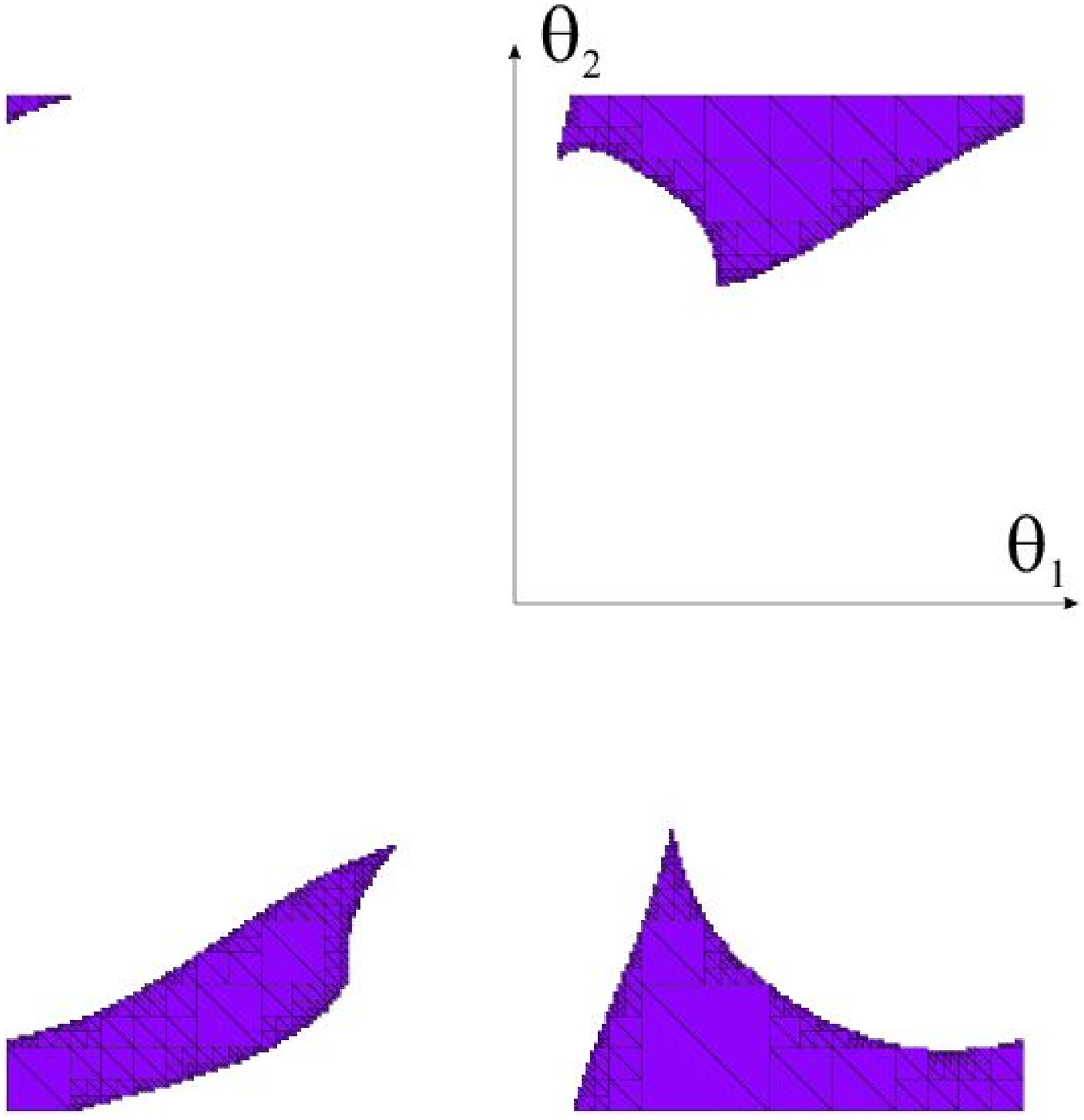, scale= 0.07}
 				\epsfig{file = 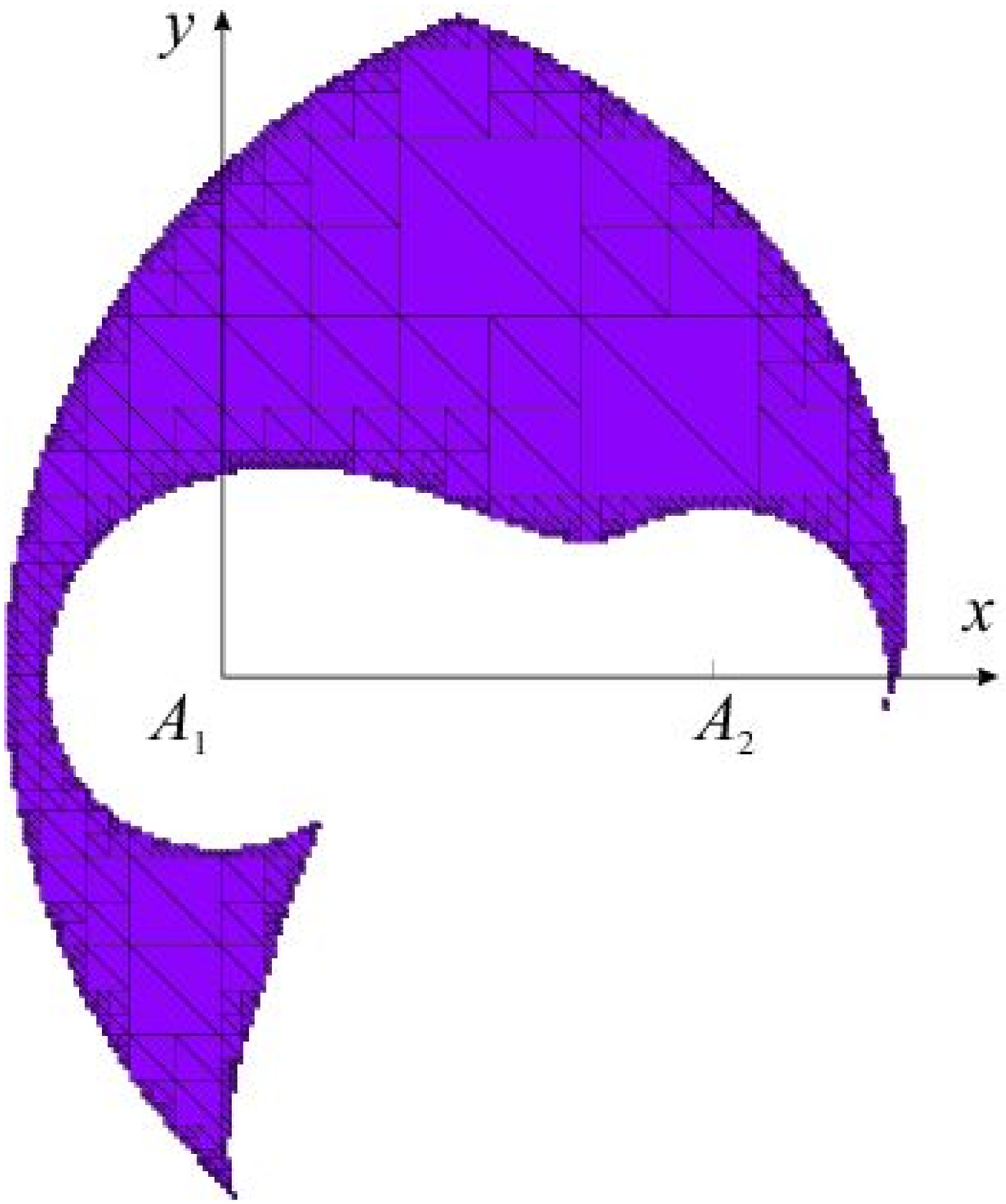, scale= 0.07}\small{(c)}
 				\epsfig{file = 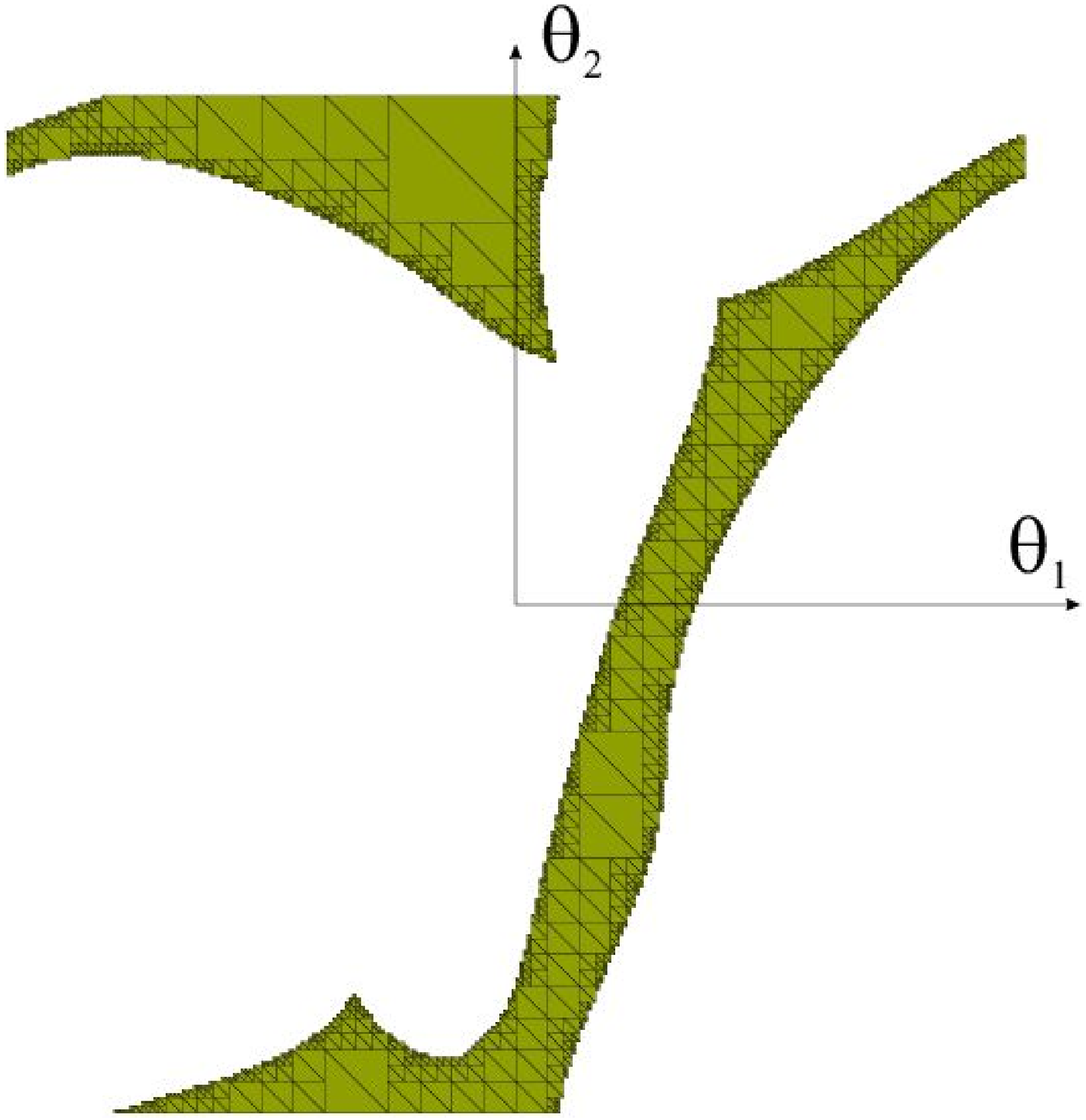, scale= 0.07}
 				\epsfig{file = 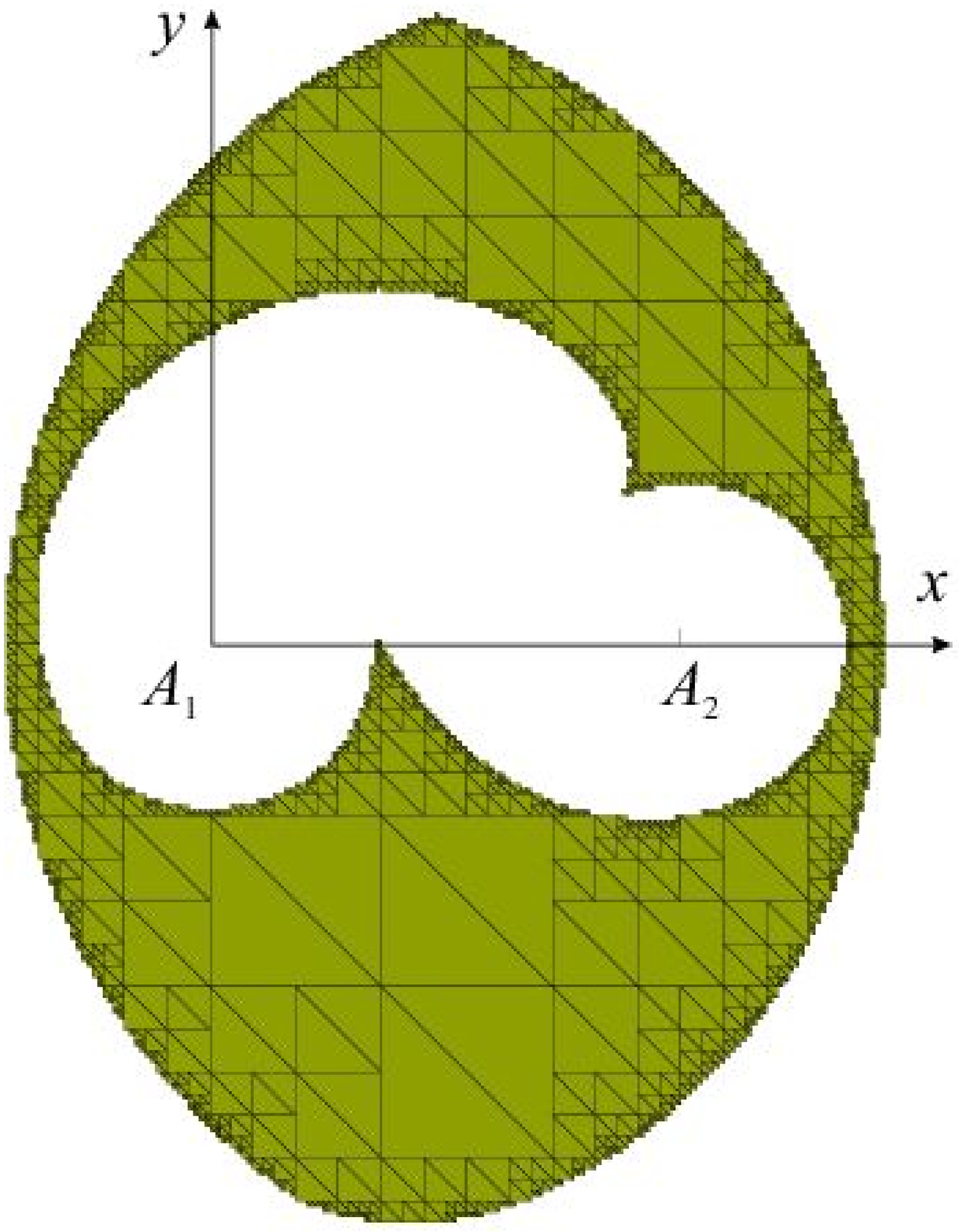, scale= 0.07}\small{(d)}
        %\vspace{-4mm}
 				\epsfig{file = 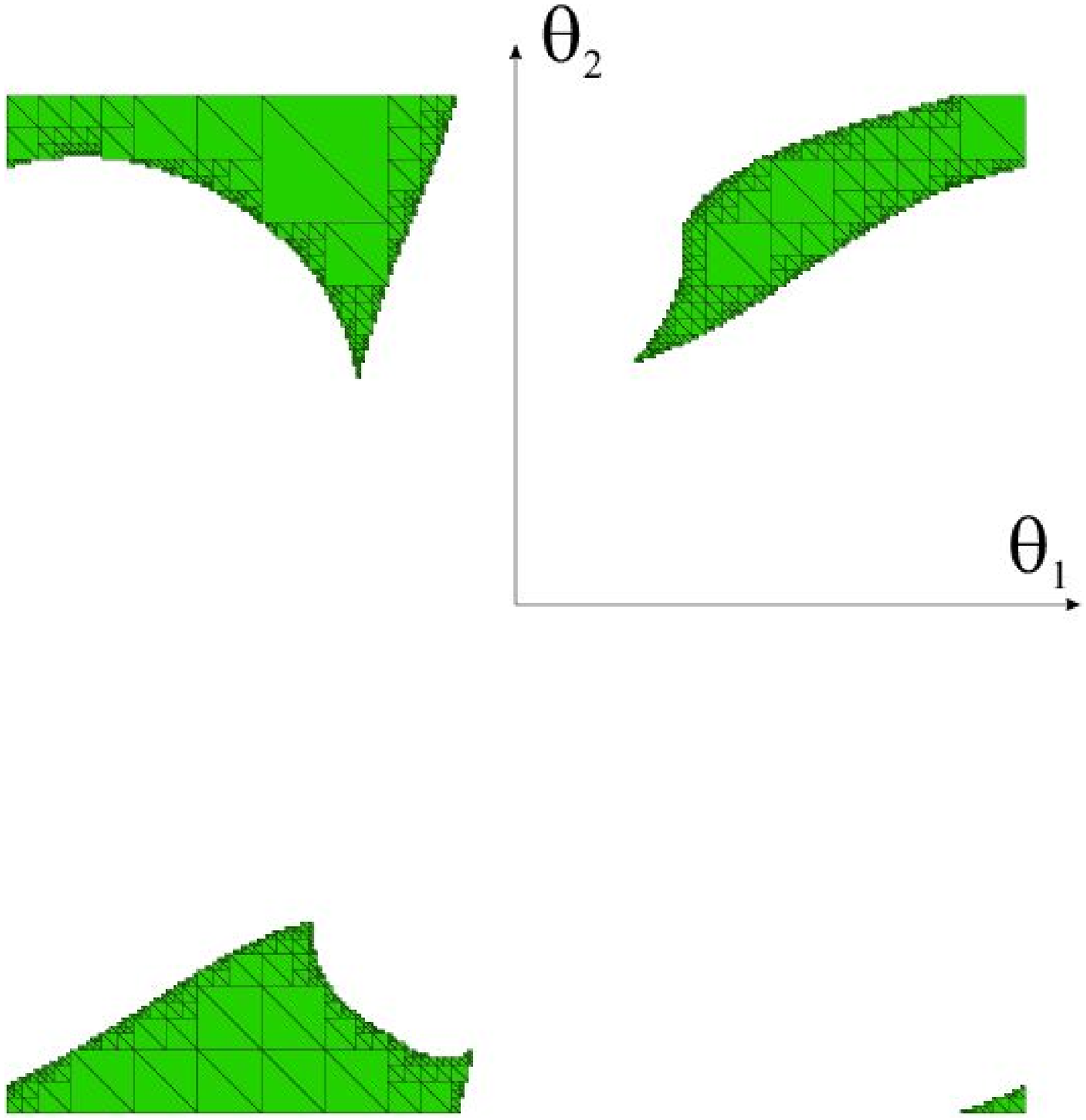, scale= 0.07}
 				\epsfig{file = 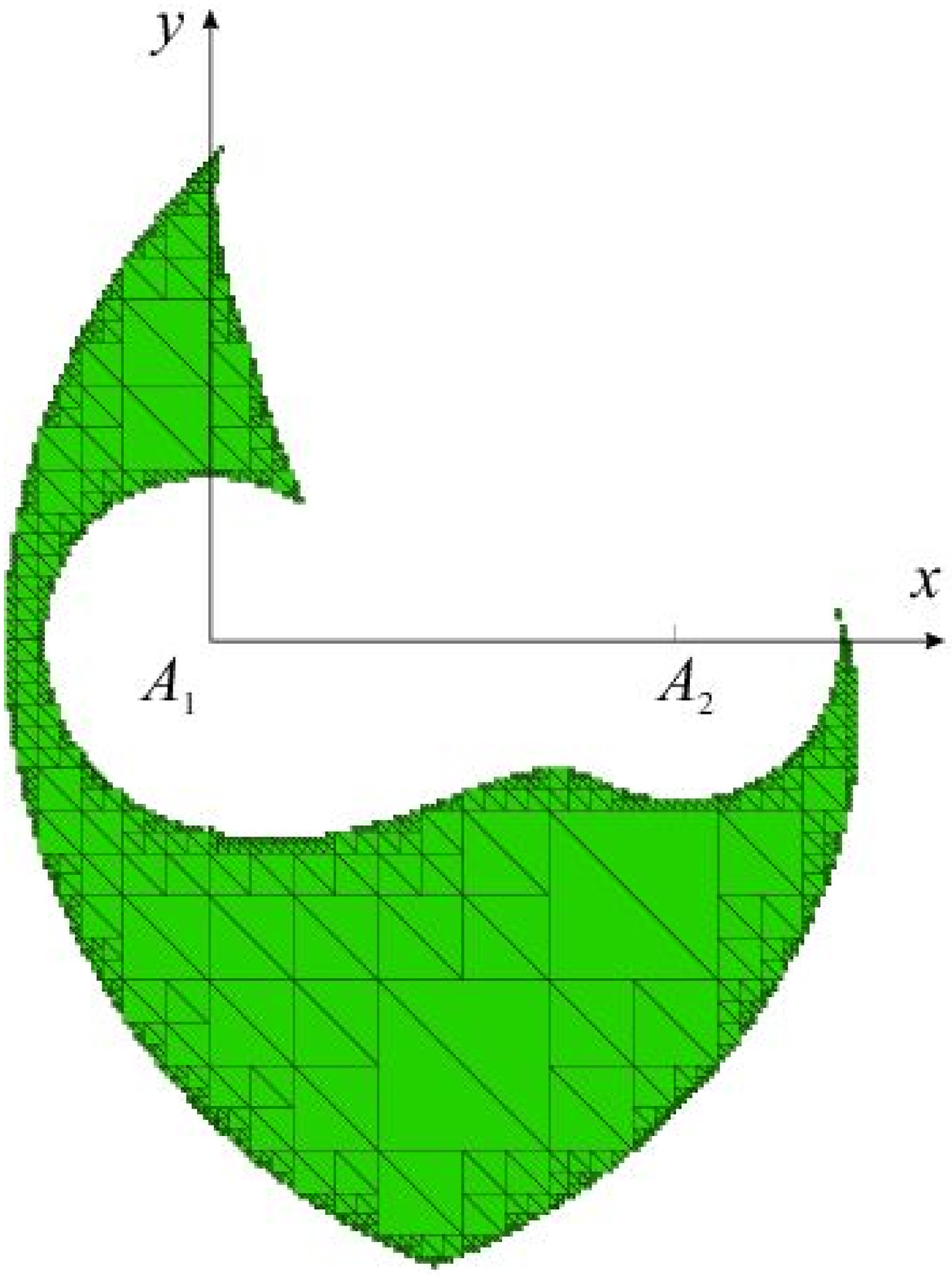, scale= 0.07}\small{(e)}
 				\epsfig{file = 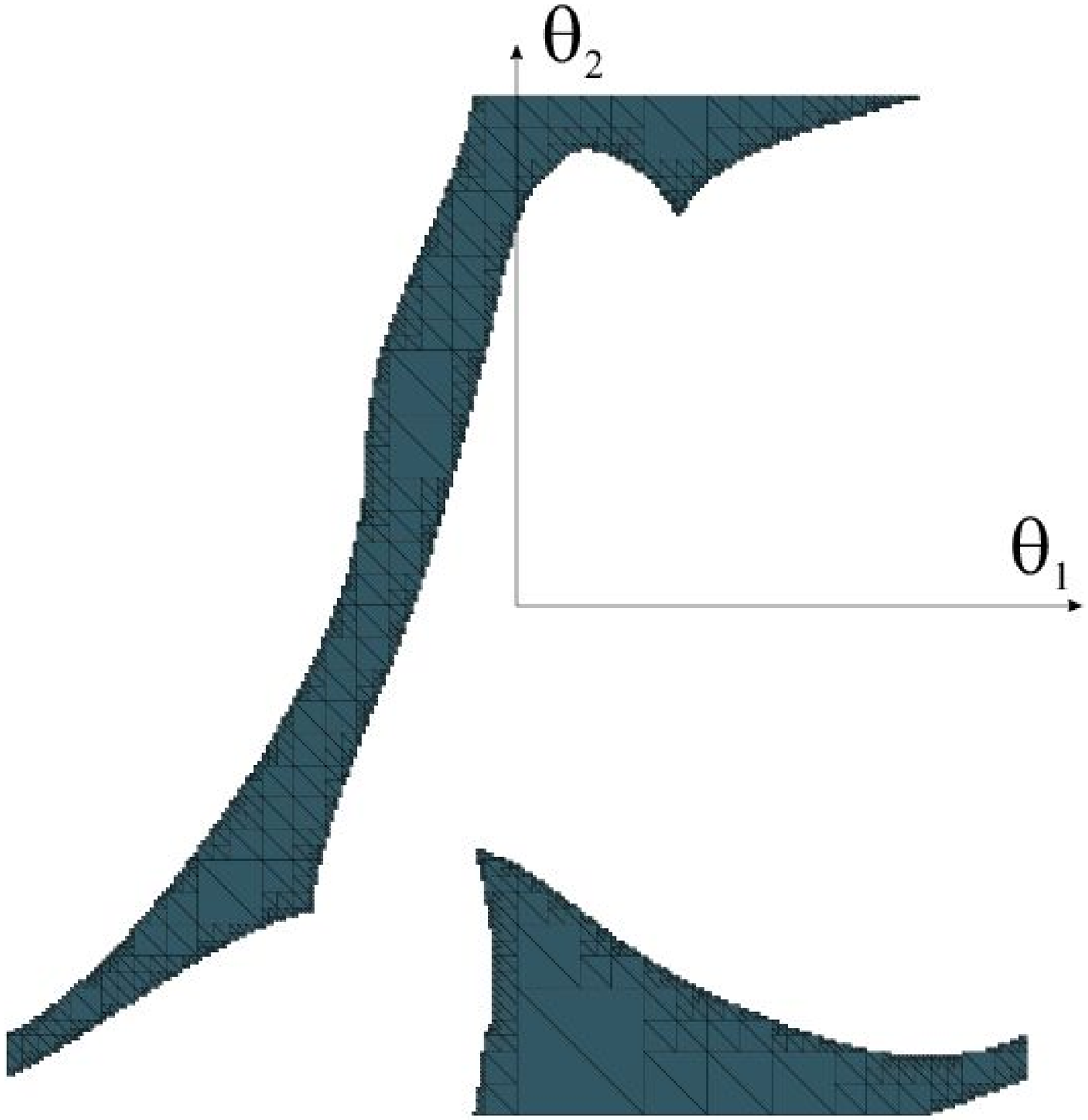, scale= 0.07}
 				\epsfig{file = 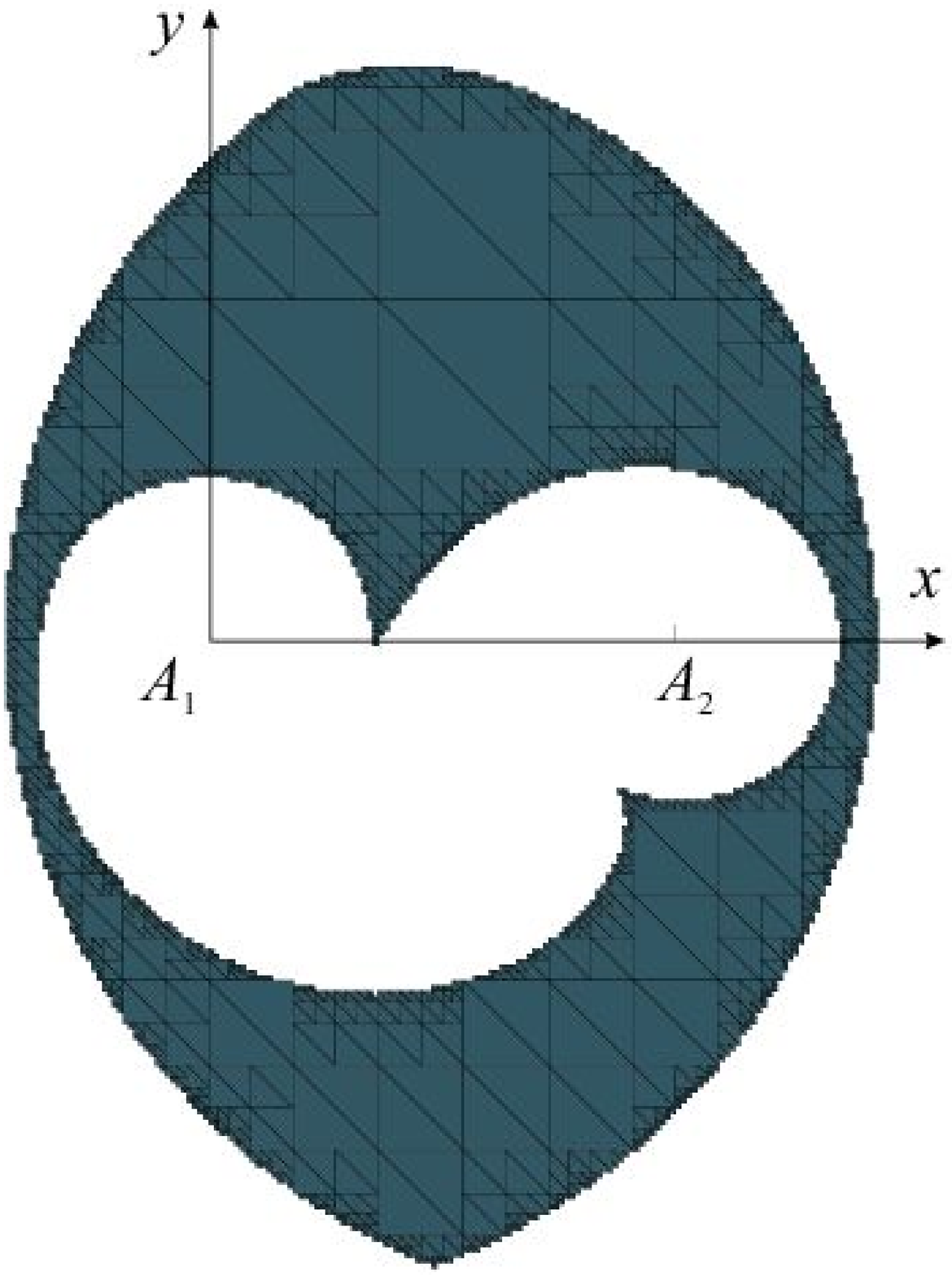, scale= 0.07}\small{(f)}
 				\epsfig{file = 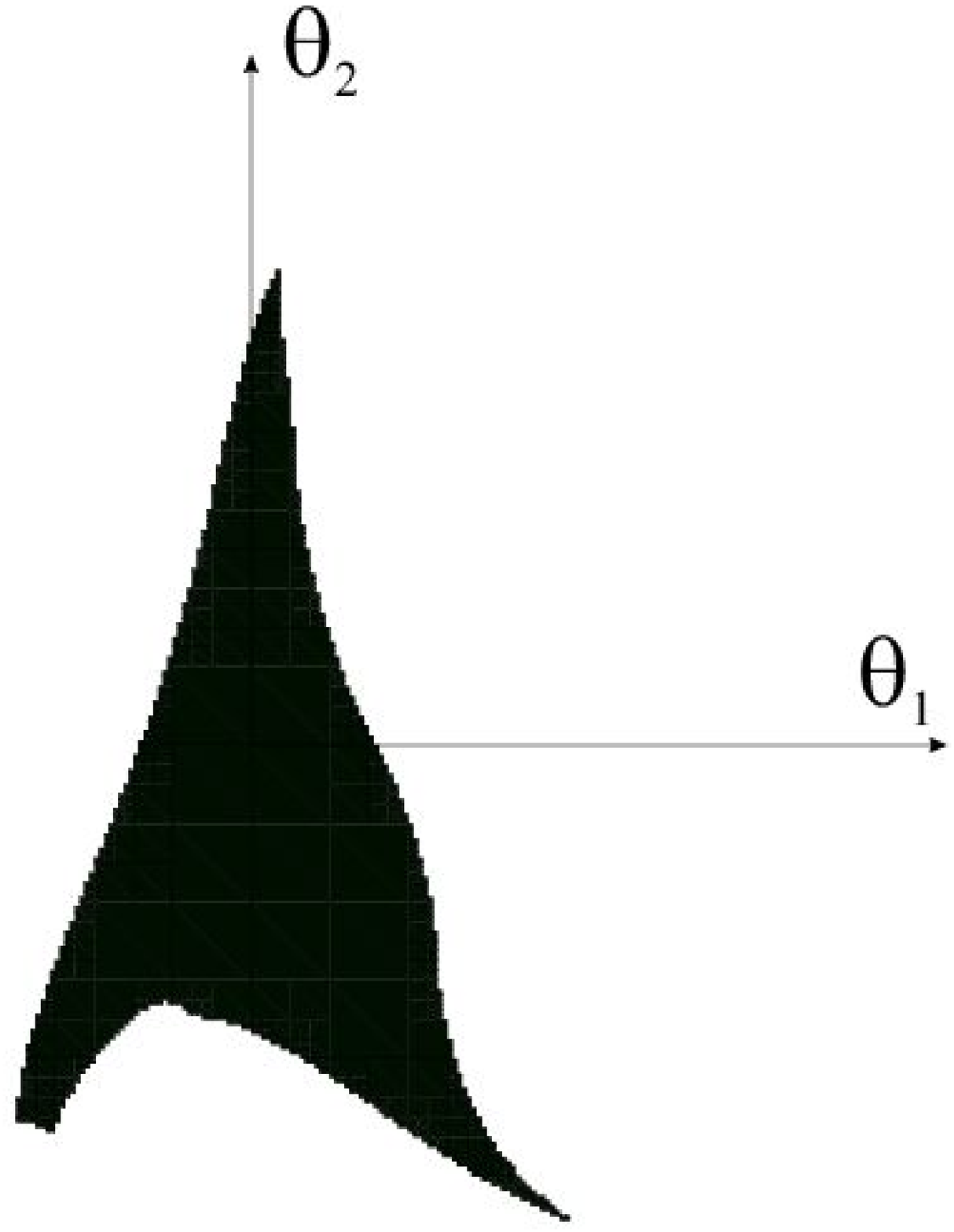, scale= 0.07}
 				\epsfig{file = 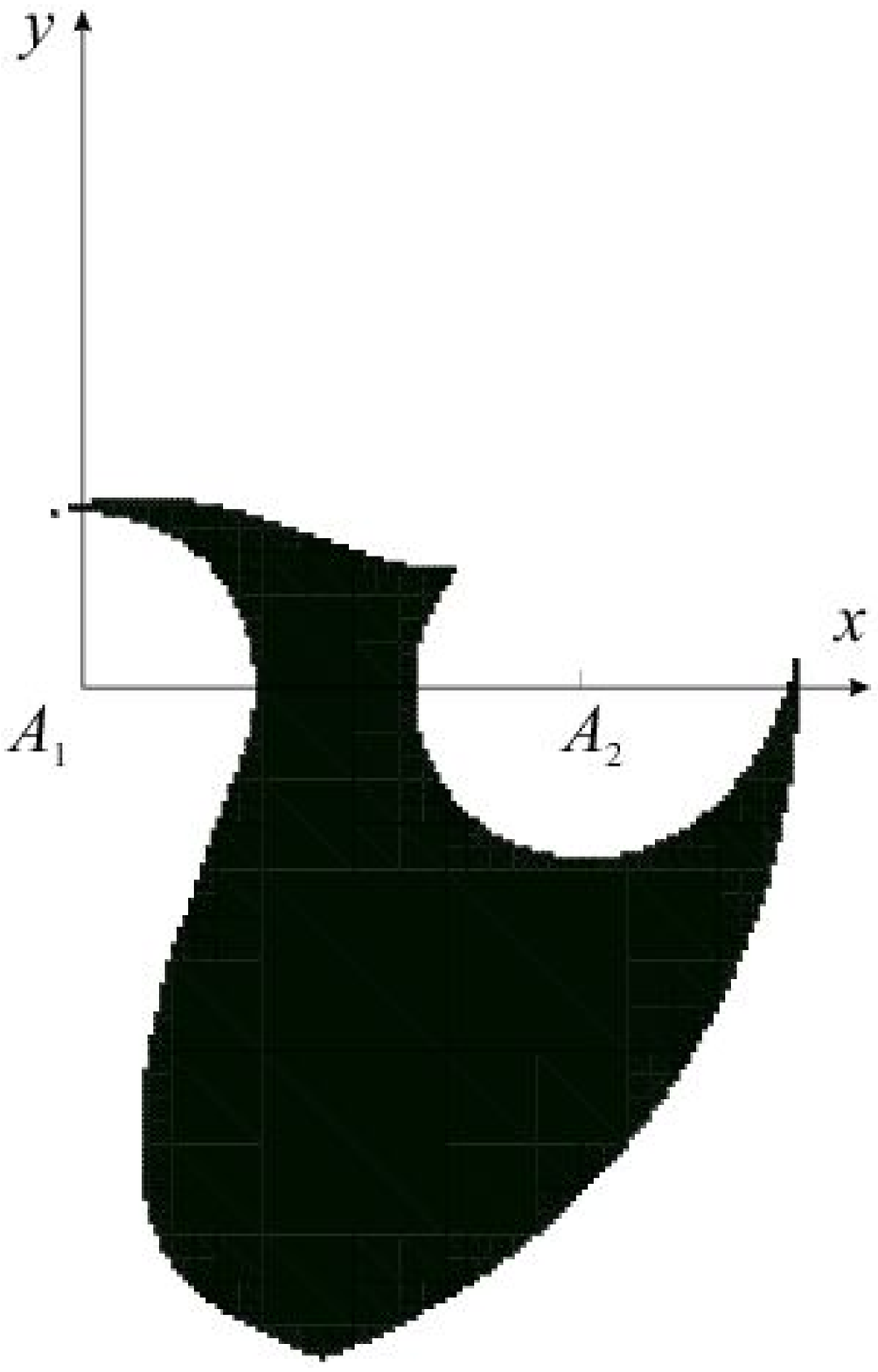, scale= 0.07}\small{(g)}
 				\epsfig{file = 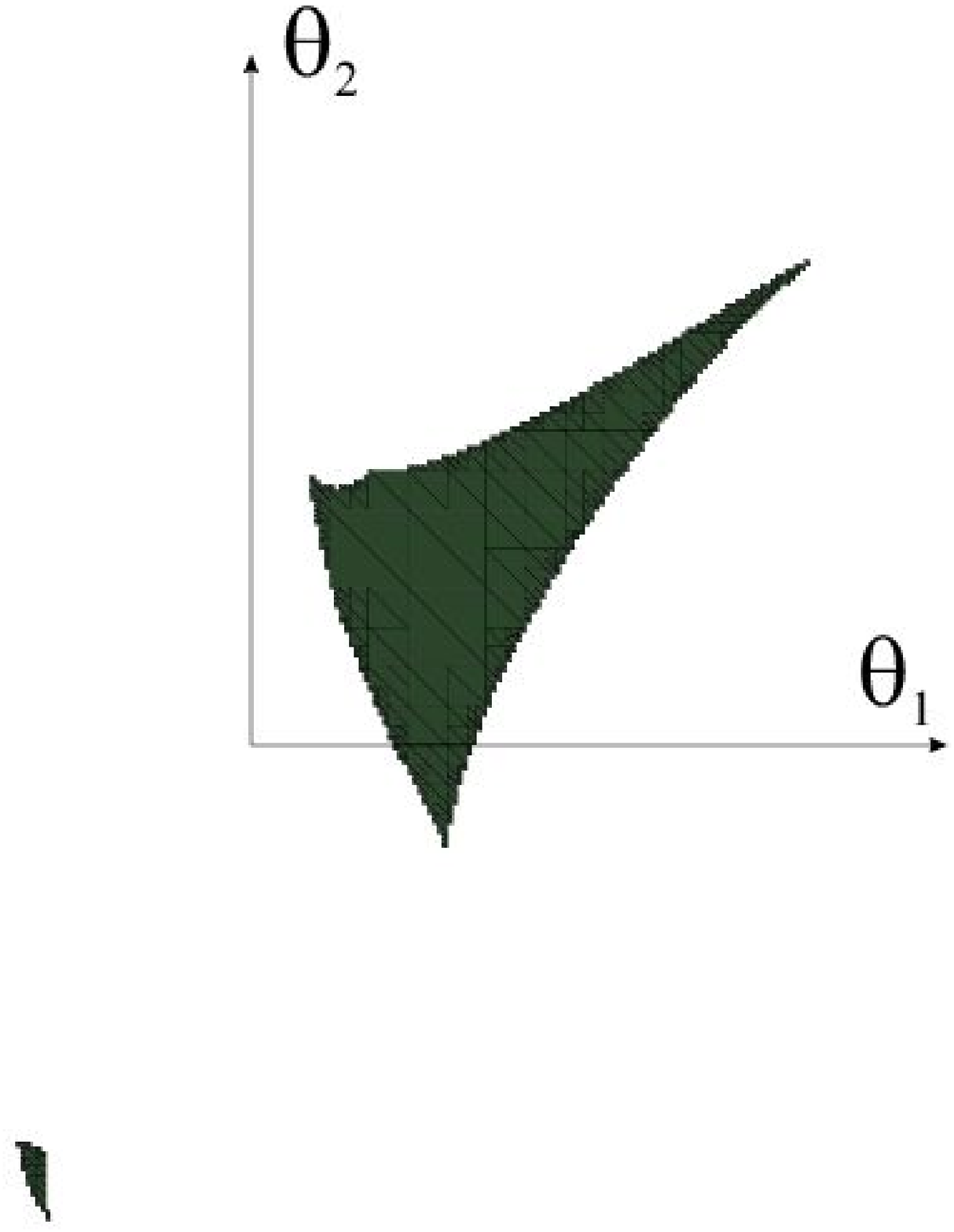, scale= 0.07}
 				\epsfig{file = 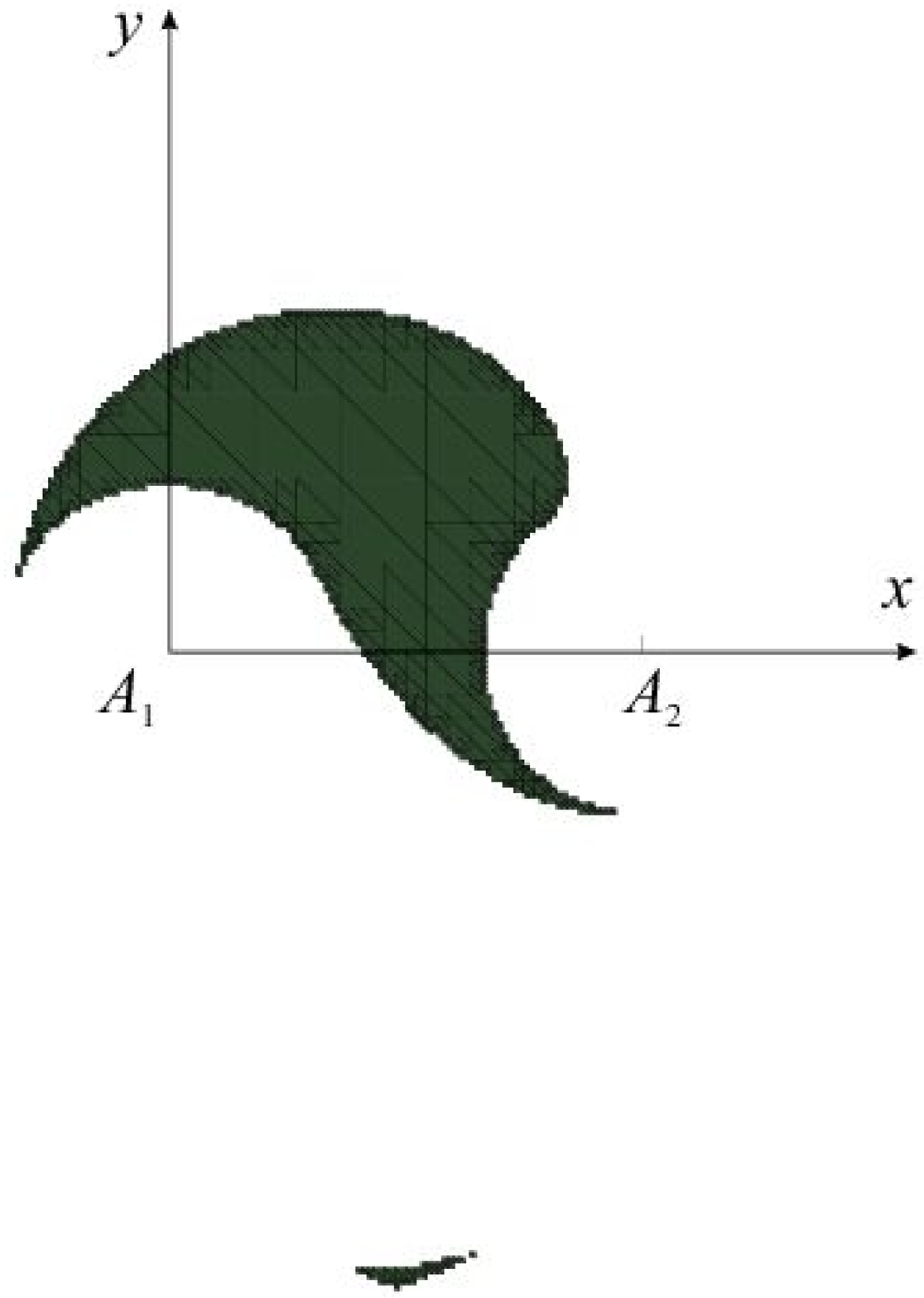, scale= 0.07}\small{(h)}
    \caption{Aspects of ${\cal M}_1$}
    \protect\label{figure:Aspect_M1}
 				\epsfig{file = 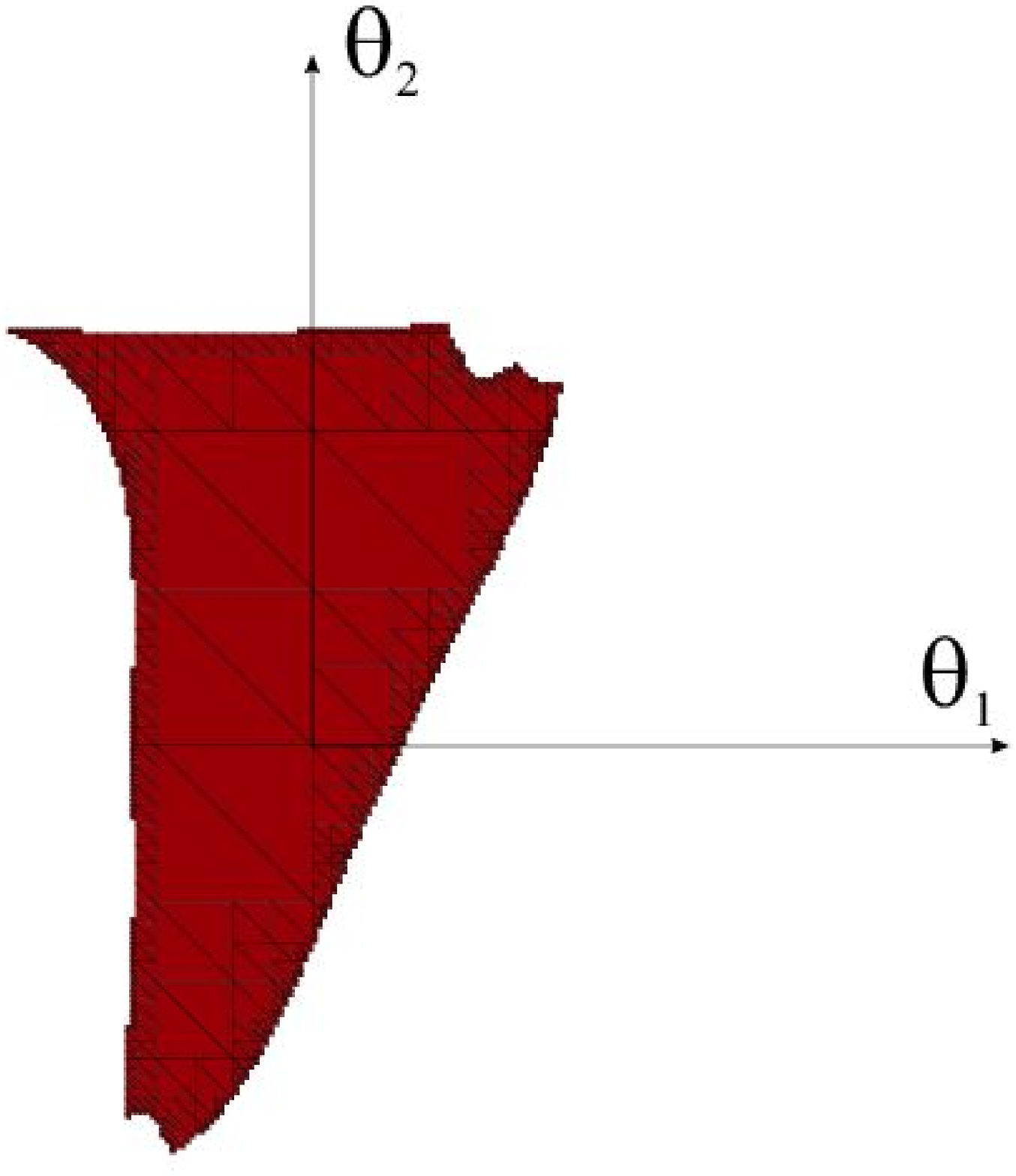, scale= 0.07}
 				\epsfig{file = 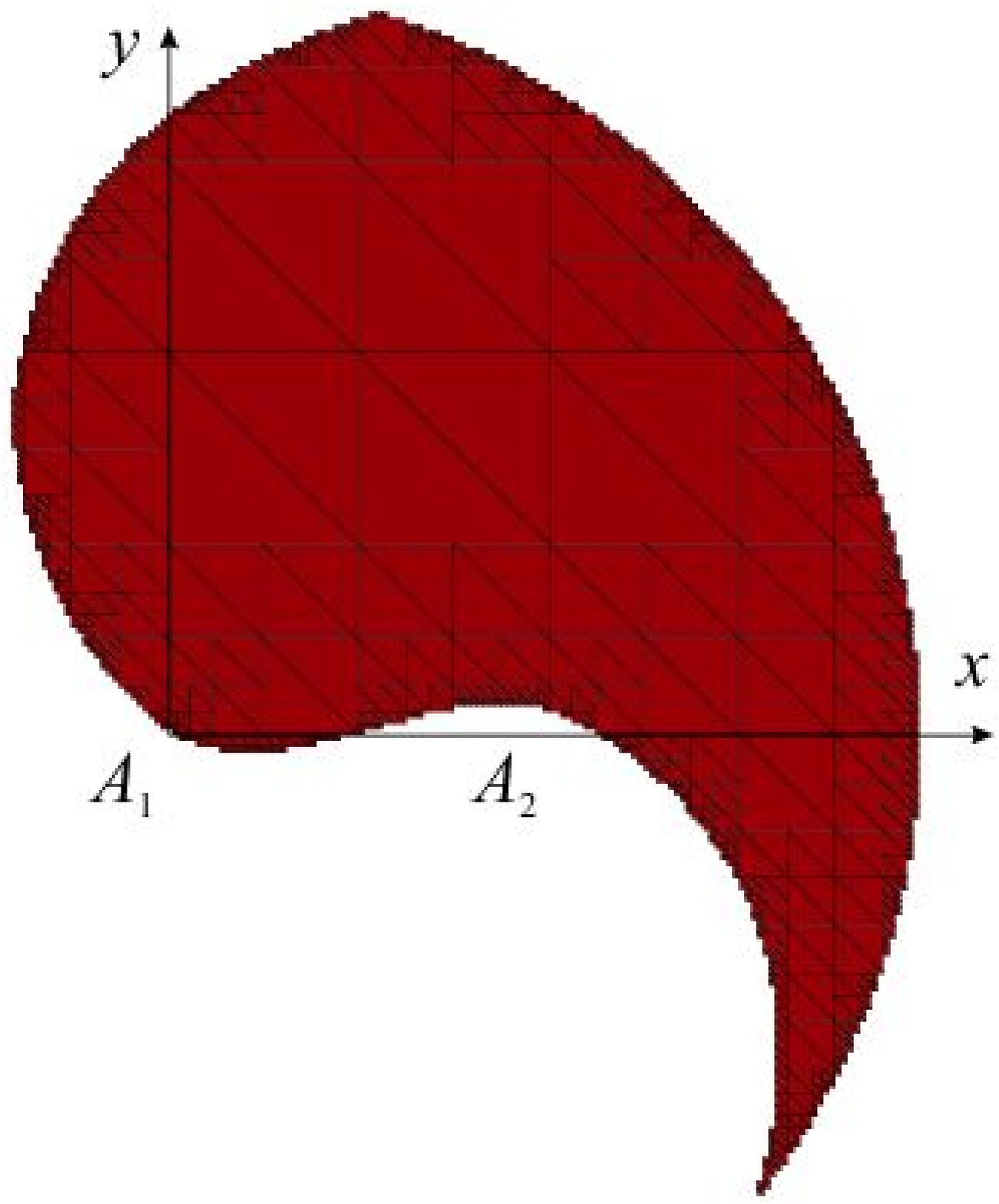, scale= 0.07}\small{(a)}
 				\epsfig{file = 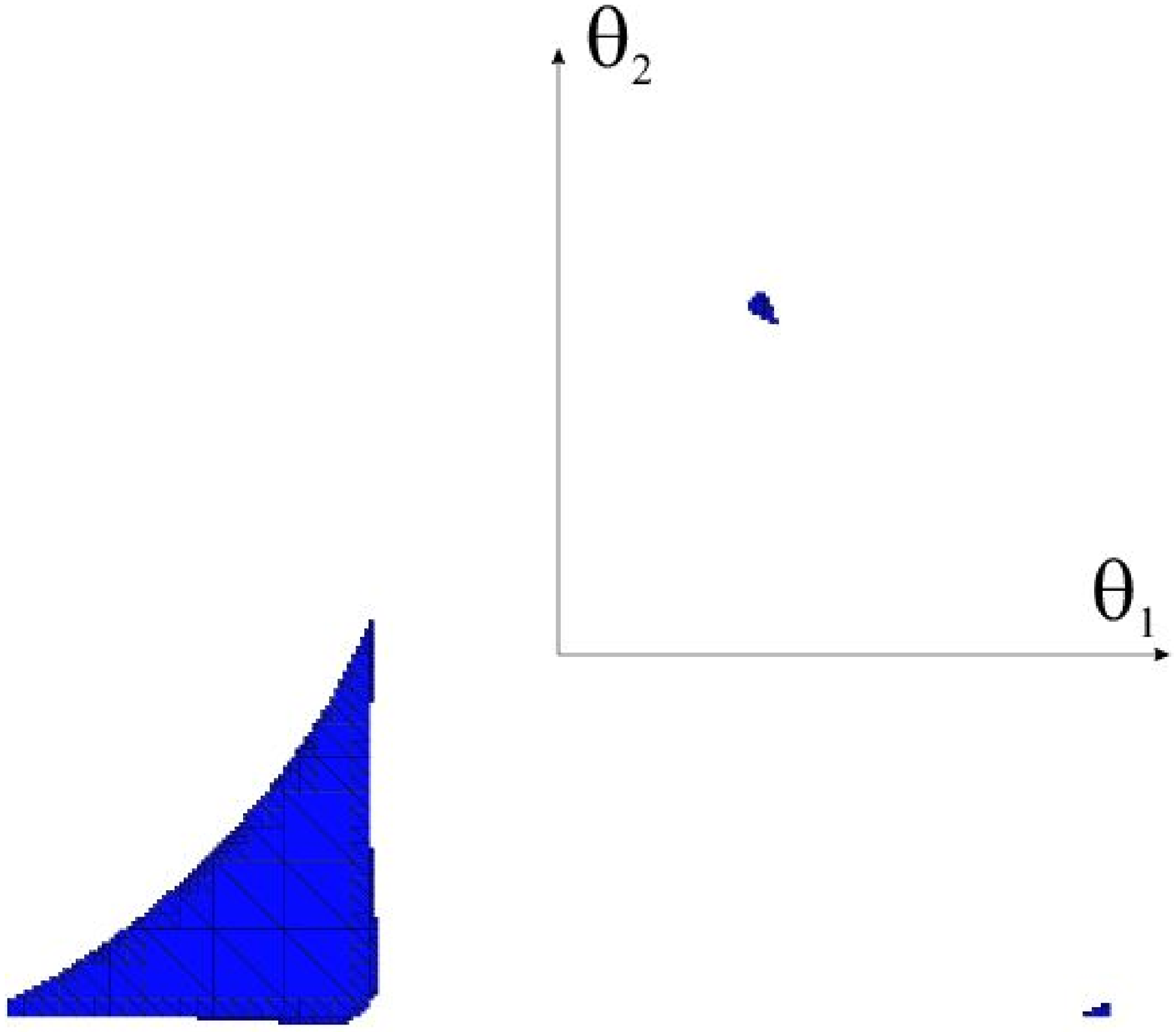, scale= 0.07}
 				\epsfig{file = 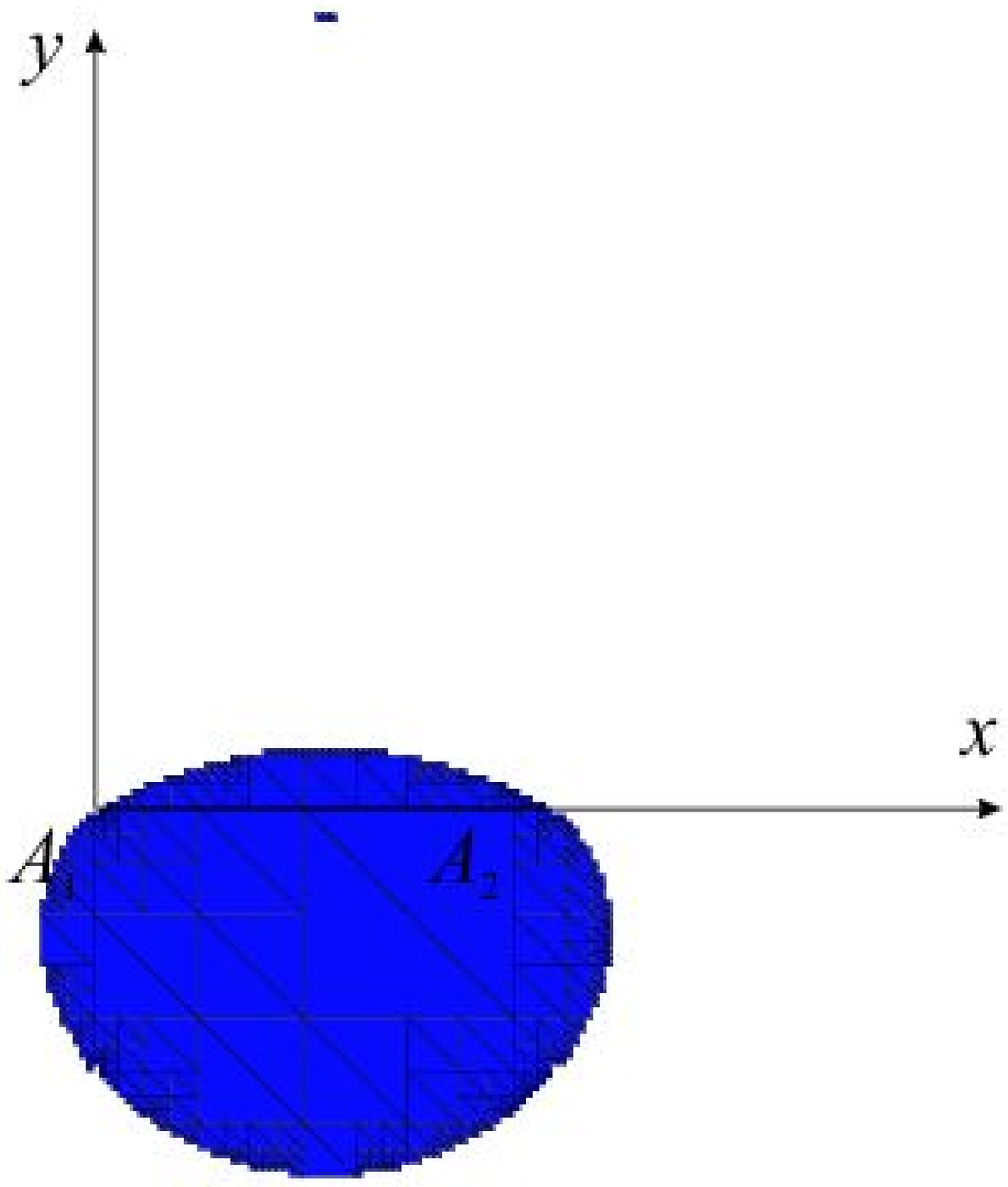, scale= 0.07}\small{(b)}
 				\epsfig{file = 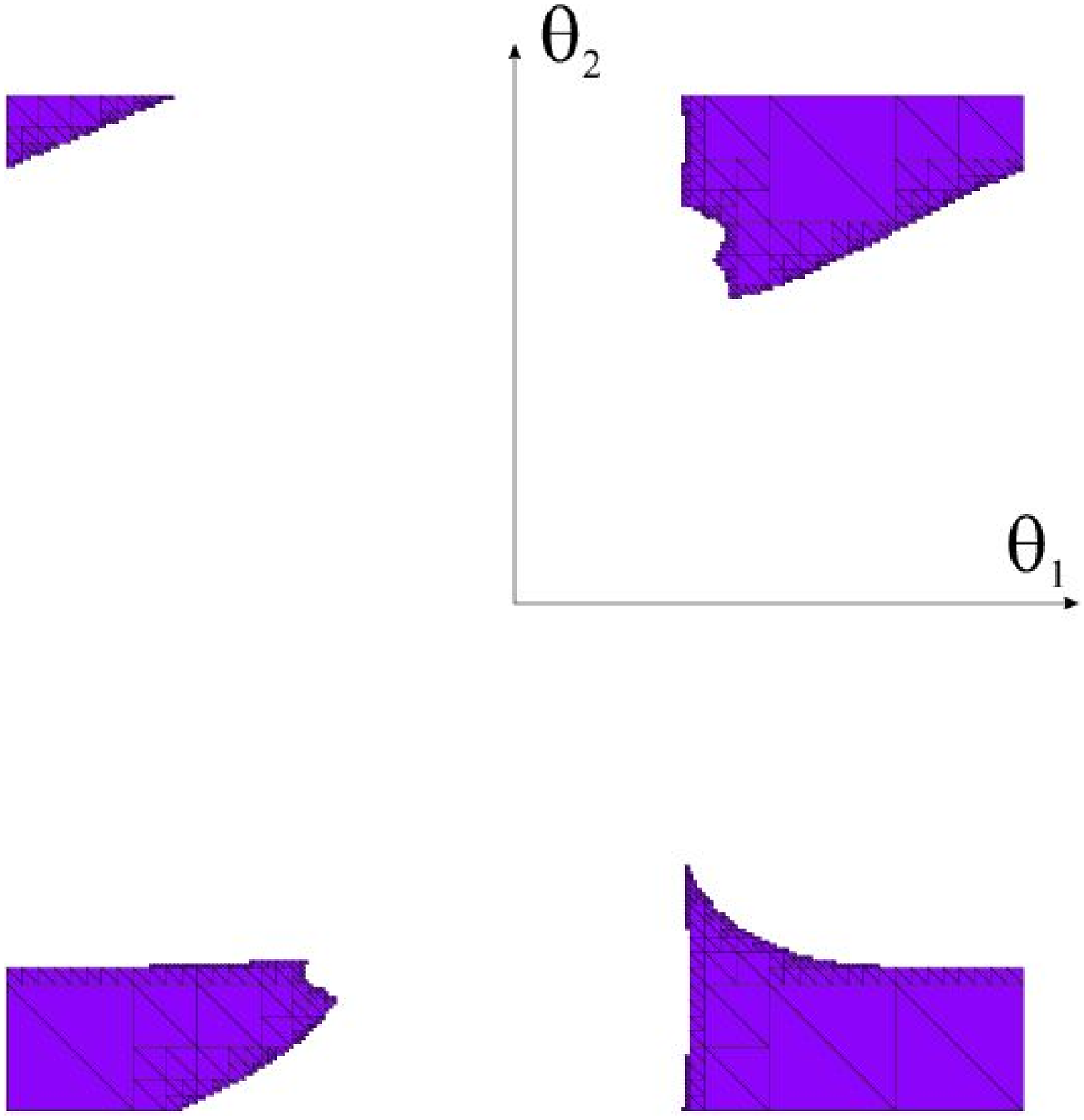, scale= 0.07}
 				\epsfig{file = 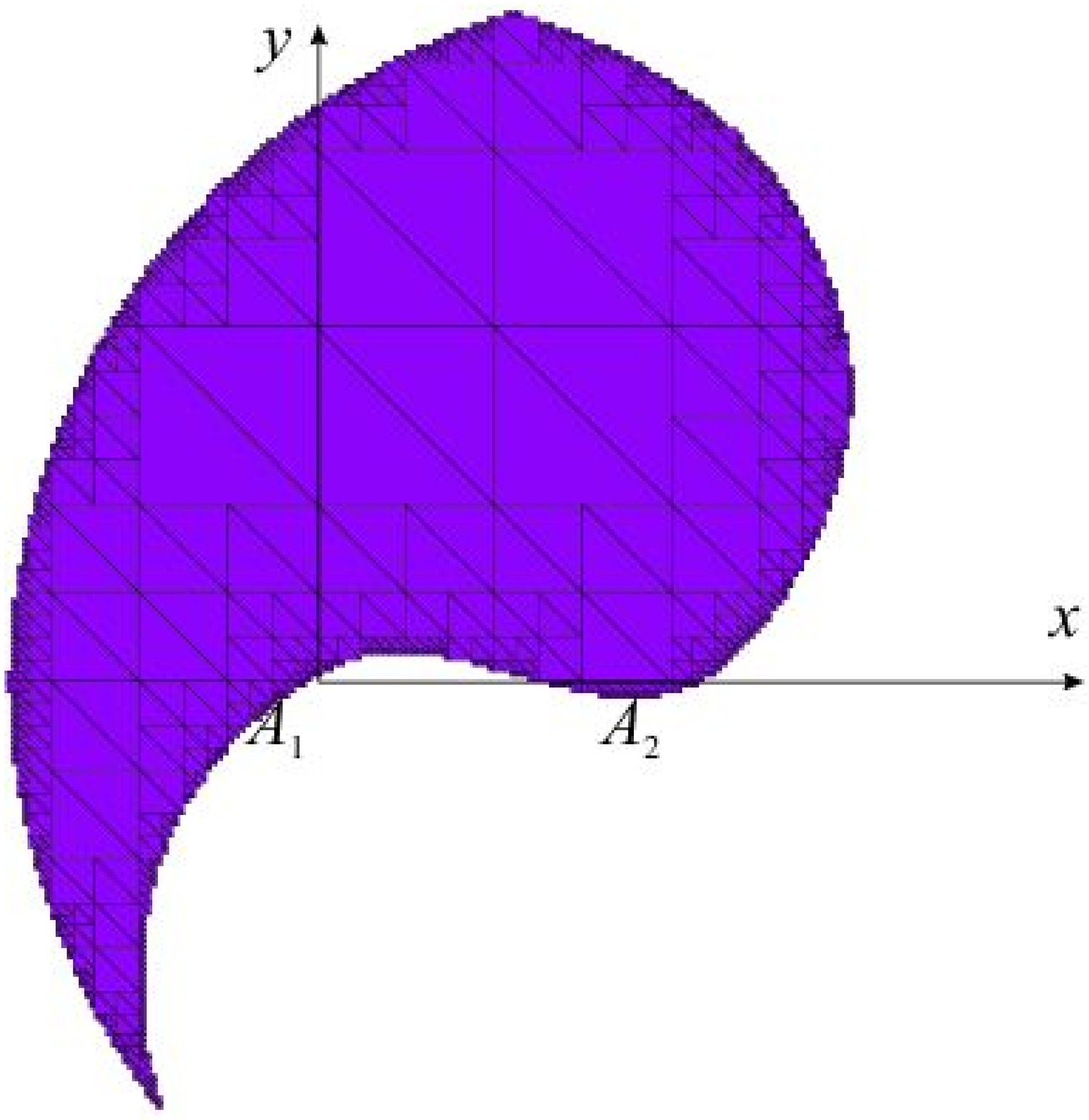, scale= 0.07}\small{(c)}
 				\epsfig{file = 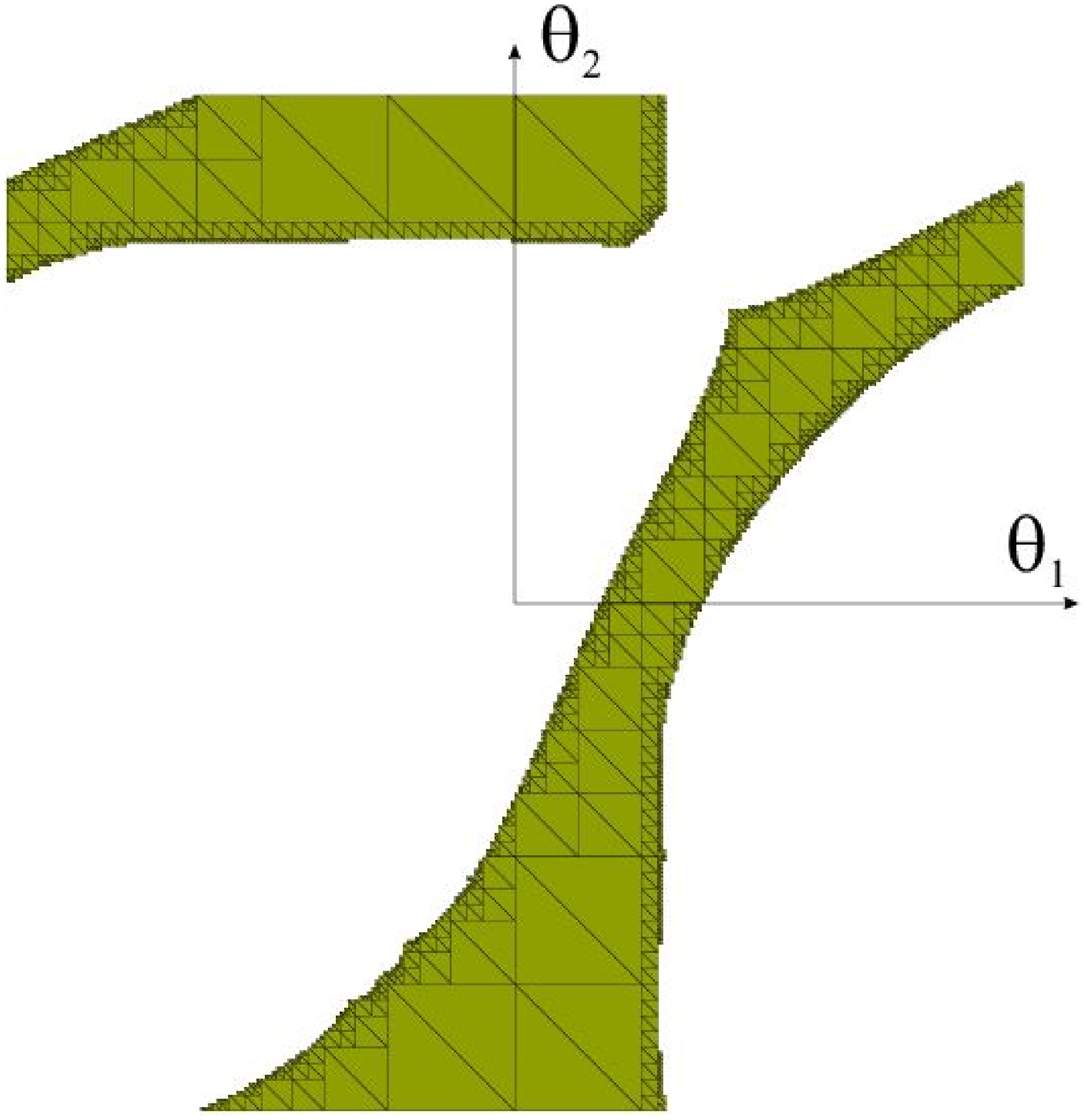, scale= 0.07}
 				\epsfig{file = 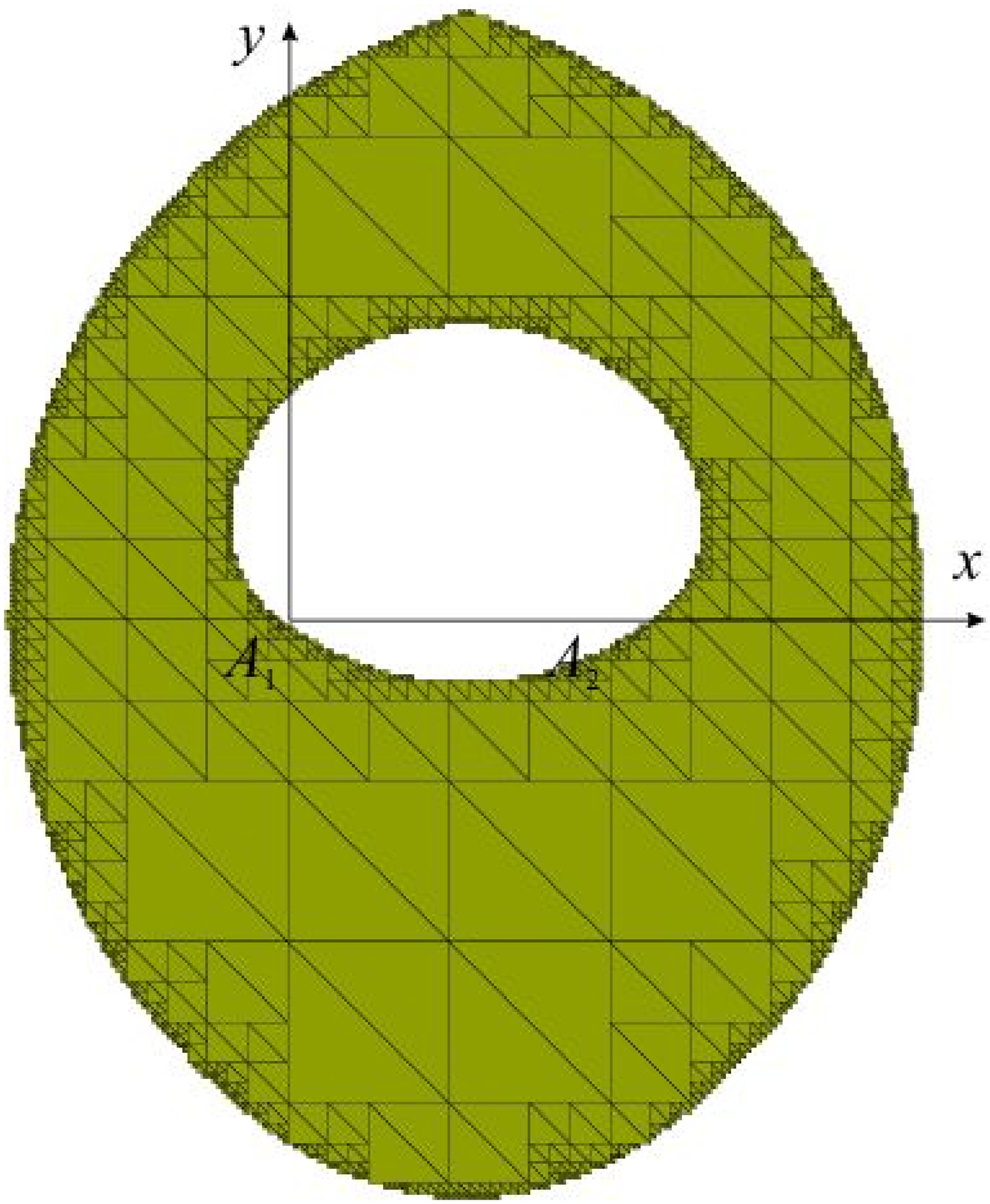, scale= 0.07}\small{(d)}
        %\vspace{-4mm}
 				\epsfig{file = 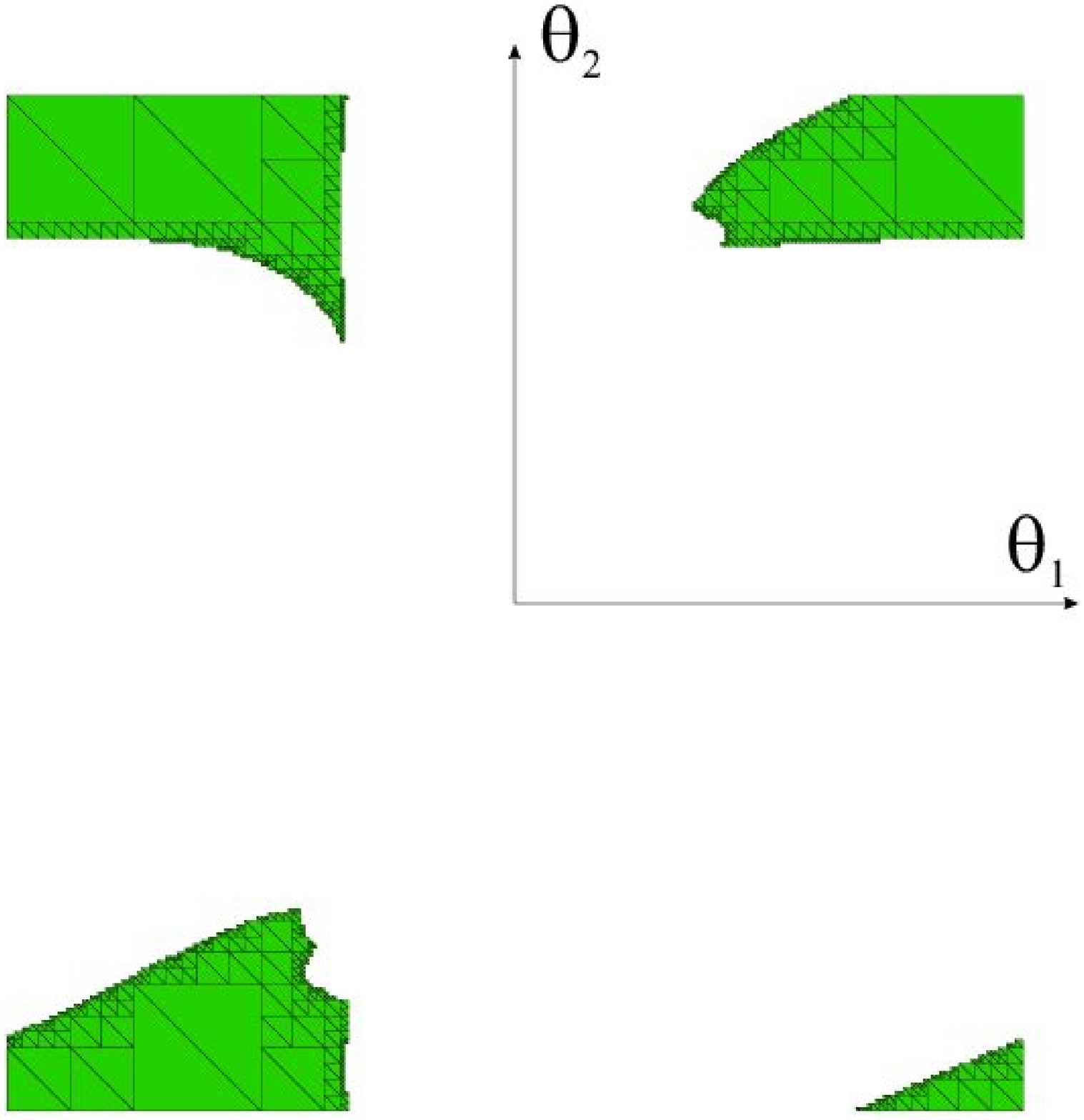, scale= 0.07}
 				\epsfig{file = 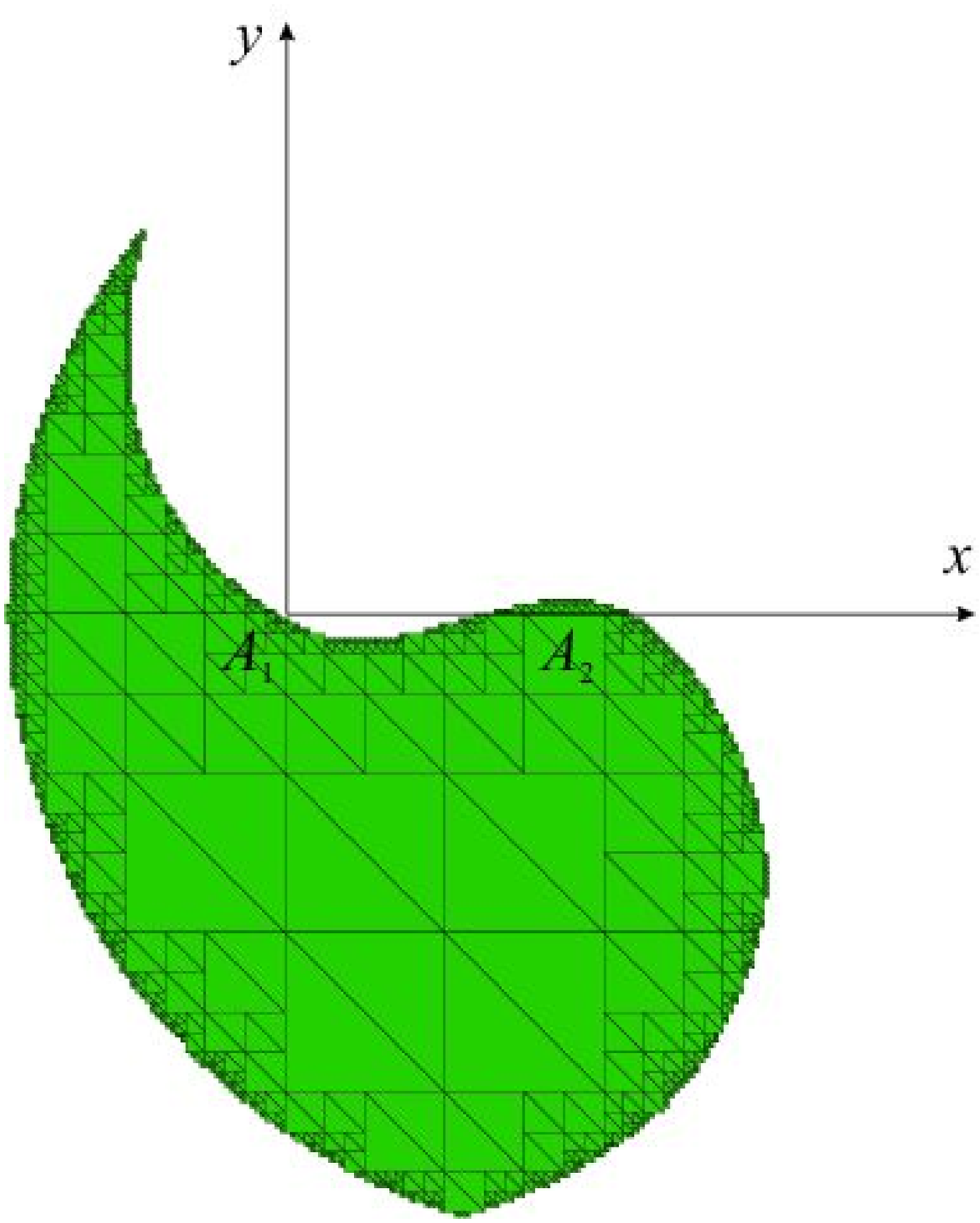, scale= 0.07}\small{(e)}
 				\epsfig{file = 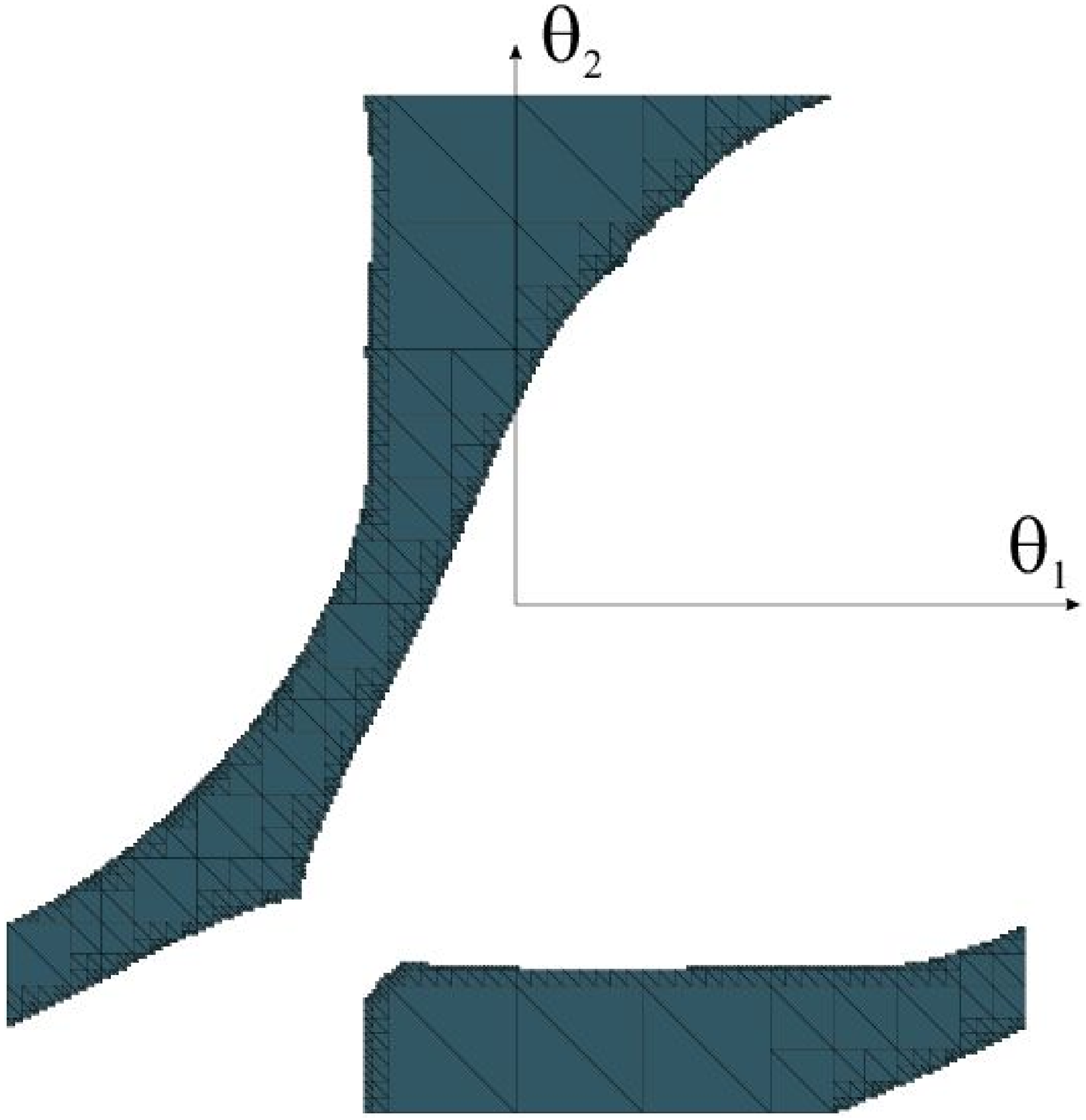, scale= 0.07}
 				\epsfig{file = 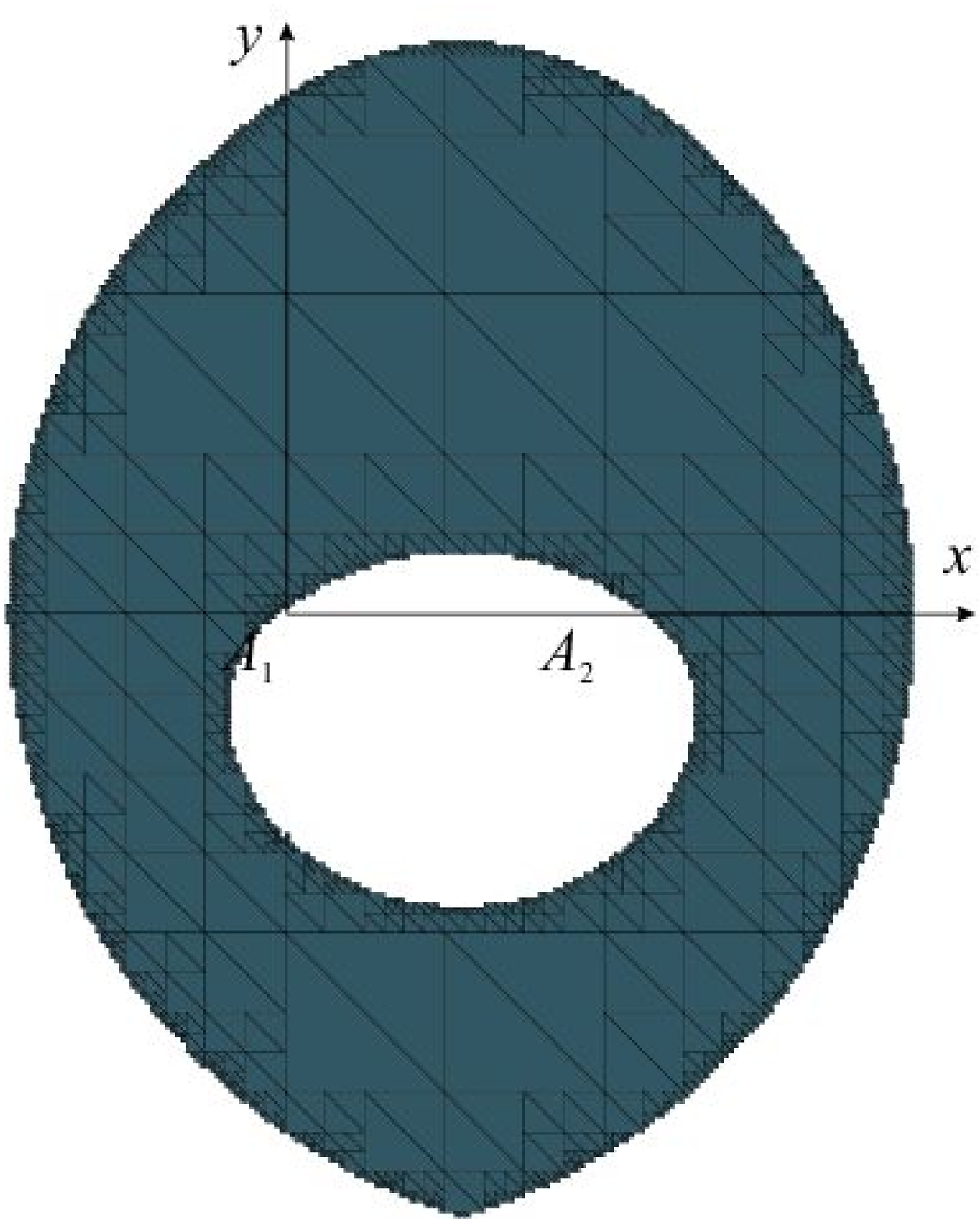, scale= 0.07}\small{(f)}
 				\epsfig{file = 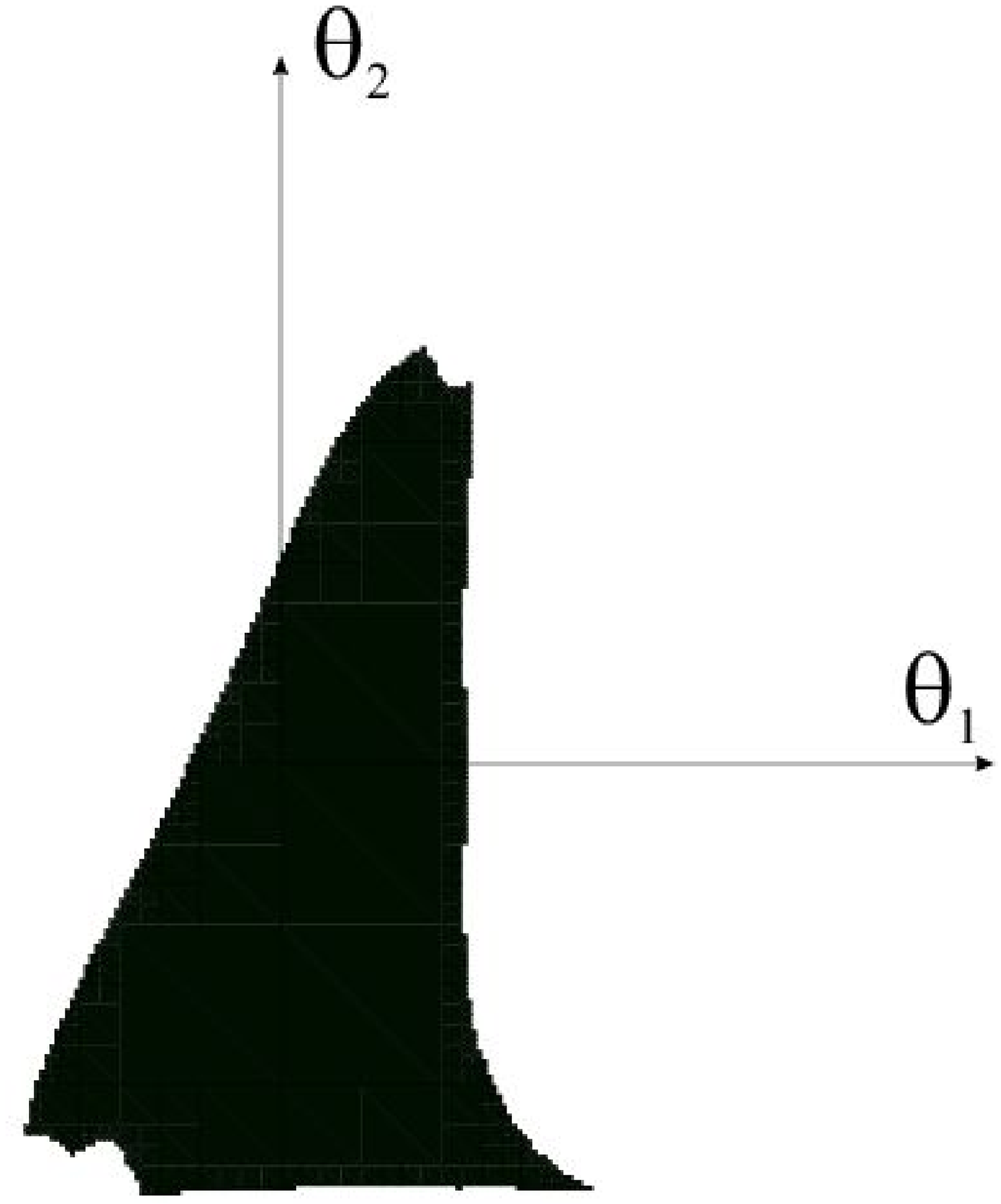, scale= 0.07}
 				\epsfig{file = 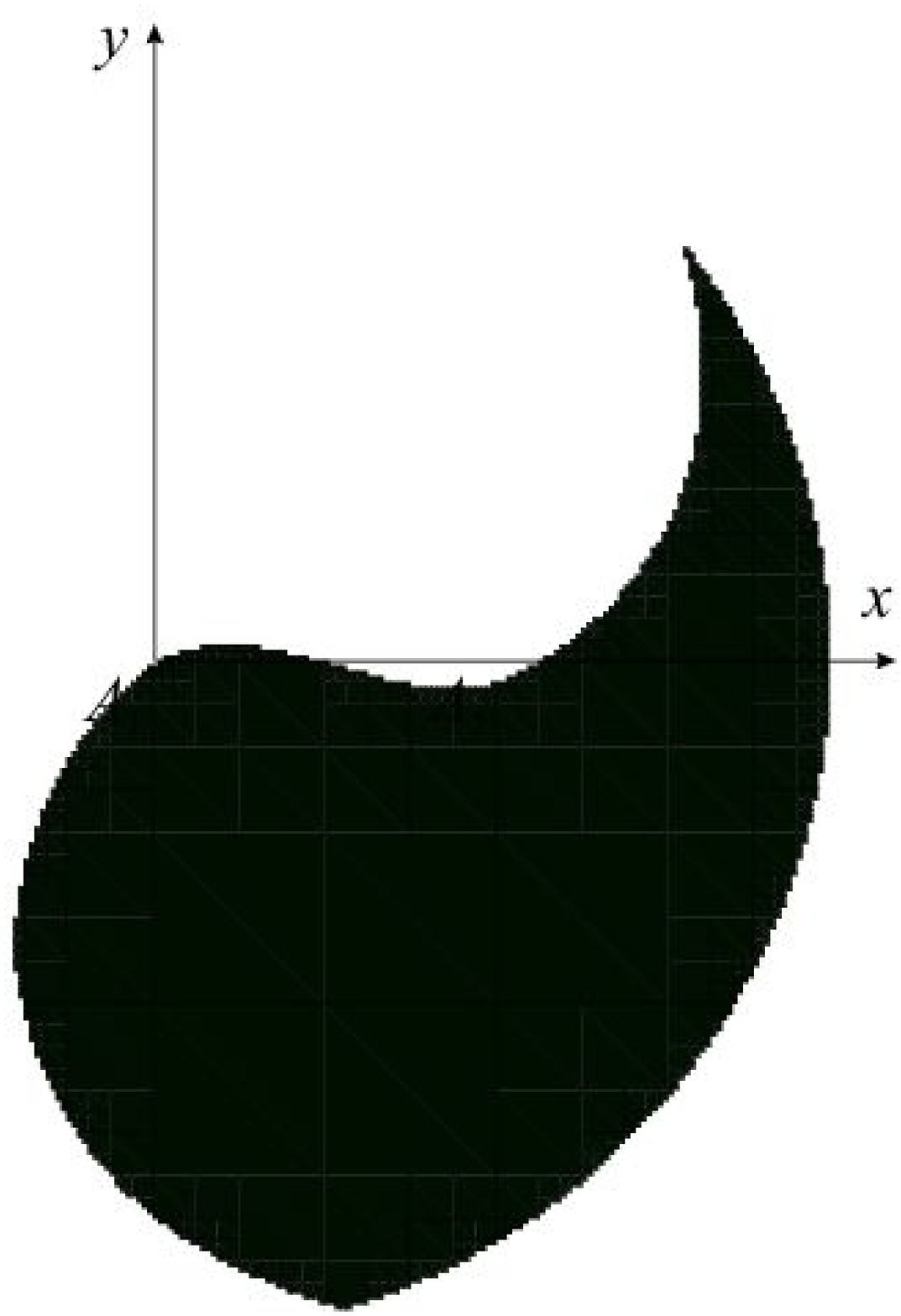, scale= 0.07}\small{(g)}
 				\epsfig{file = 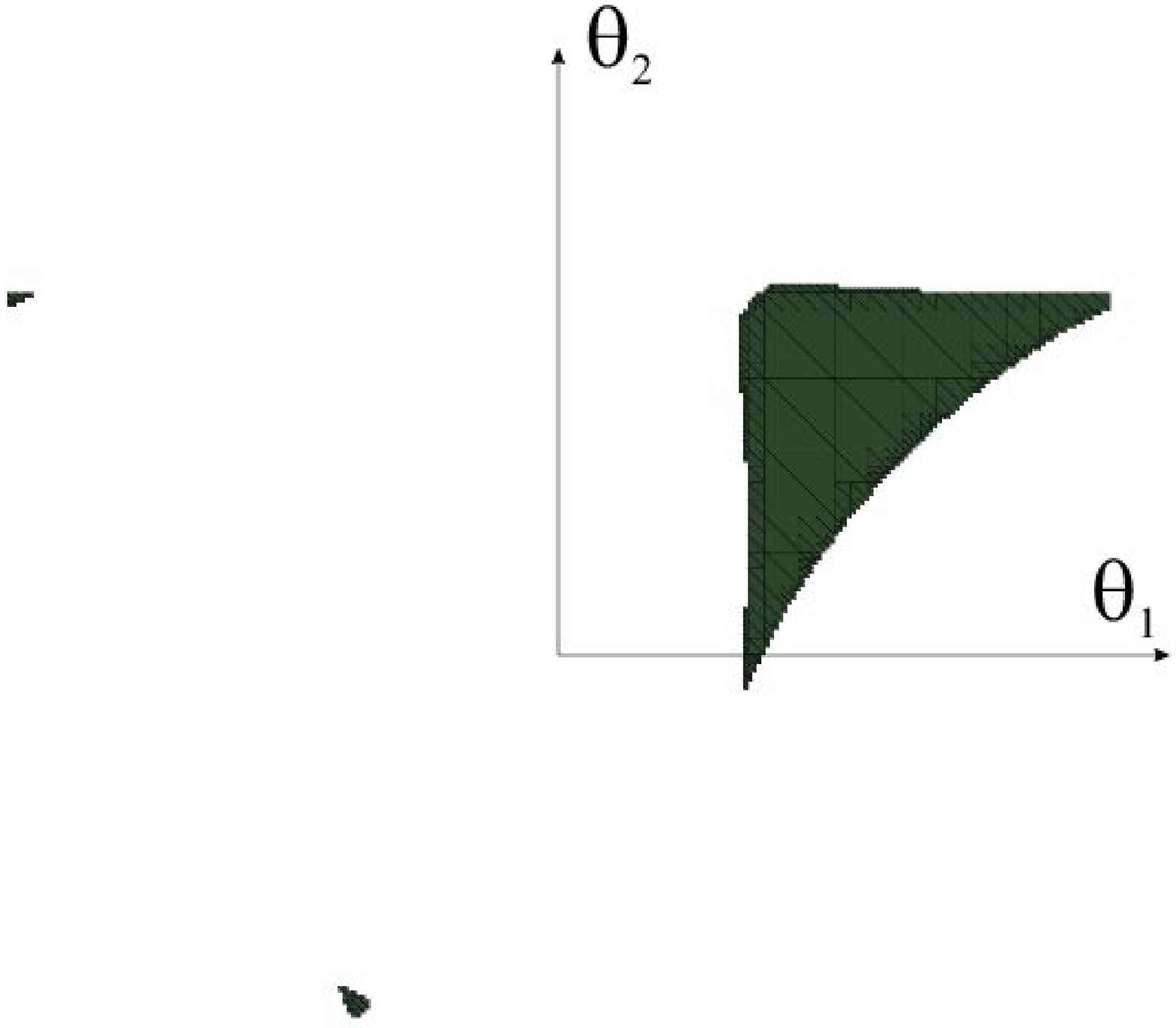, scale= 0.07}
 				\epsfig{file = 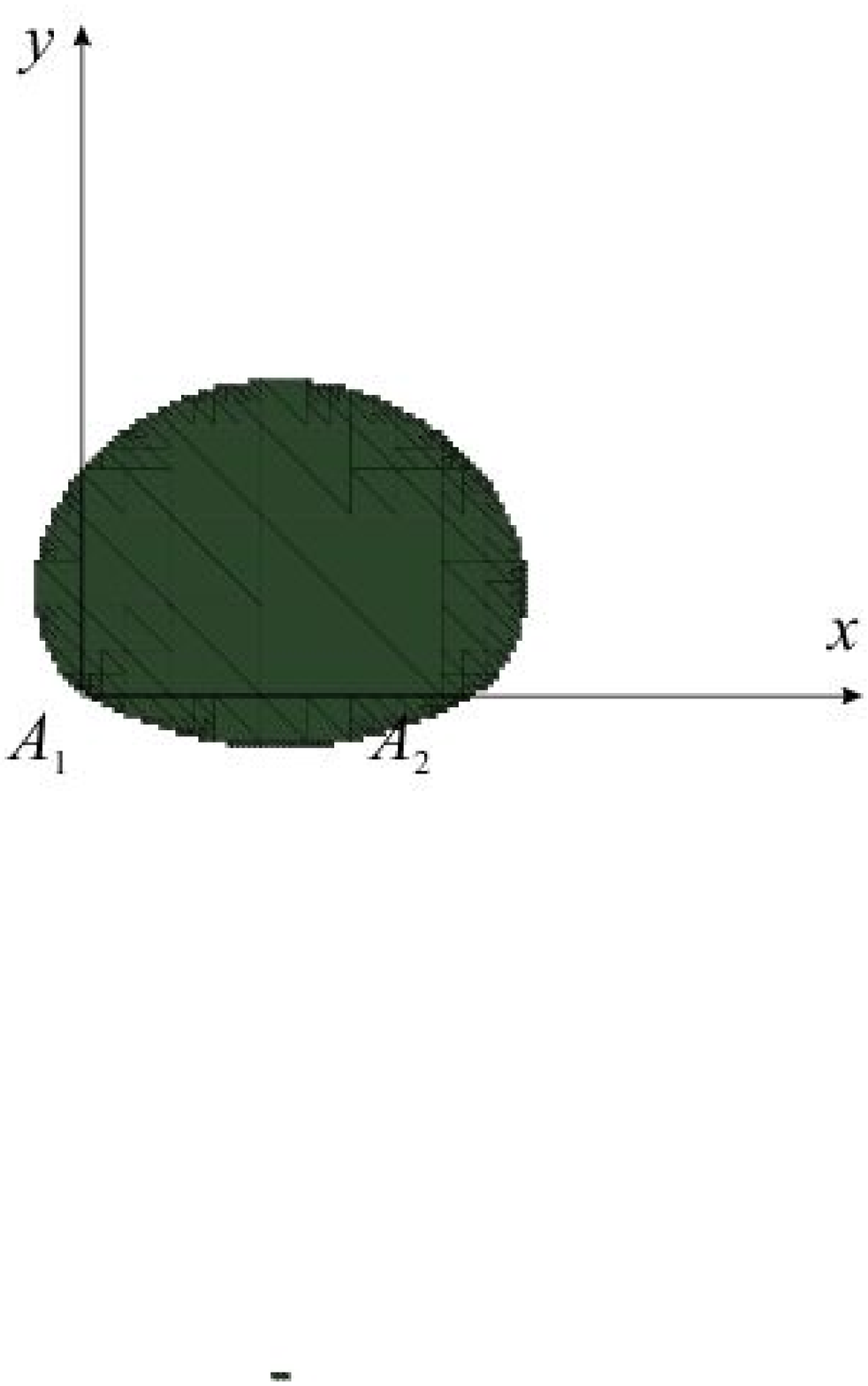, scale= 0.07}\small{(h)}
    \caption{Aspects of ${\cal M}_2$}
    \protect\label{figure:Aspect_M2}
    \end{center}
\end{figure}
%%%%%%%%%%%%%%%%%%%%%%%%%%%%%%%%%%%%%%%%%%%%%%%%%%%%%%%%%%%%%%%%%%%%%%%%%%%%%%%%
\subsection{Comparison between classical quadtree computation and the new algorithm}
%%%%%%%%%%%%%%%%%%%%%%%%%%%%%%%%%%%%%%%%%%%%%%%%%%%%%%%%%%%%%%%%%%%%%%%%%%%%%%%%
The aim of this section is to compare the number of times where the inverse or direct kinematic function is called to build the quadtree for the both mechanisms and for several  depths of the tree which is equivalent to the accuracy of the model. For the joint space, the initial box $B$ is defined by two intervals equal to $[-\pi, \pi]$ and for the workspace, the initial box $B$ is defined by two intervals equal to $[-(L_1+L_3), (L_1+L_3)]$.

With the discretization method, the number of times where the inverse or direct kinematic problem is used is $n_{discretization}= 2^{2d}$ with $d$ is the depth of the tree.  We call $n_{quadtree}$ the number of times where where the inverse or direct kinematic problem is used to build the quadtree model. To compare the computing cost between the both method, we define the following criteria:
\be
  {\cal K} = n_{quadtree} / n_{discretization}
\ee
Table~\ref{table_example_jointspace_workspace} compares the computing times for the joint space and the workspace for the mechanisms ${\cal M}_1$ and ${\cal M}_2$. When the depth of the quadtree is small, there is no advantage to used the interval analysis based method. But, when the depth increase, the advantages can be very important. For example, to compute the workspace of ${\cal M}_1$ with $d=10$, we call 36893 times the IKP while the discretization method calls $2^{20}=1048576$ times the IKP.
\begin{table}
   \begin{center}
      \caption{Comparison between classical quadtree computation and the new algorithm to compute the joint space and the workspace}
      \label{table_example_jointspace_workspace}
      \begin{tabular}{|c|c|c|c|c|c|c|c|}
      \hline
      Joint space & Depth  \# & 5 & 6 & 7 & 8 & 9 & 10\\ \hline
       & ${\cal M}_1$ &  99\% & 73\% & 40\% & 22\% & 13 \% & 9 \% \\
       & ${\cal M}_2$ & 111\% & 83\% & 40\% & 18\% &  8 \% & 4 \% \\
      \hline
      Workspace &  Depth  \# & 5 & 6 & 7 & 8 & 9 & 10\\ \hline
       &  ${\cal M}_1$ &  72\% & 45\% & 25\% & 13\% &  7 \% & 3 \% \\
       &  ${\cal M}_2$ &  65\% & 37\% & 19\% & 10\% &  5 \% & 2 \% \\
      \hline
      \end{tabular}
   \end{center}
\end{table}
%%%%%%%%%%%%%%%%%%%%%%%%%%%%%%%%%%%%%%%%%%%%%%%%%%%%%%%%%%%%%%%%%%%%%%%%%%%%%%%%
\section{Conclusions and future works}
%%%%%%%%%%%%%%%%%%%%%%%%%%%%%%%%%%%%%%%%%%%%%%%%%%%%%%%%%%%%%%%%%%%%%%%%%%%%%%%%
In this paper, we have presented an algorithm able to compute the joint space and workspace of parallel mechanism by using the octree model and the interval based method. We have the maximum singularity regions, called {\it aspects}, for two planar mechanisms and we have compared the number of times where the inverse and direct kinematic problem is called according to the accuracy used. Thanks to the use of the interval analysis based method, the result is guaranteed for all the depth used, i.e. we are sure to detect all the singular configurations. The quadtree model can be saved in text file and its size is very small. Plugin exist now to visualize the model in a 3D viewer in Web pages (as www.octree.com).  Future works will be made to define as interval the lengths of the legs and to find the influence of the manufacturing error on the joint space and workspace.
%%%%%%%%%%%%%%%%%%%%%%%%%%%%%%%%%%%%%%%%%%%%%%%%%%%%%%%%%%%%%%%%%%%%%%%%%%%%%%%%
\section{Acknowledgments}
%%%%%%%%%%%%%%%%%%%%%%%%%%%%%%%%%%%%%%%%%%%%%%%%%%%%%%%%%%%%%%%%%%%%%%%%%%%%%%%%
This work was supported partly by the French Research Agency A.N.R. (Agence Nationale pour la Recherche). 
%%%%%%%%%%%%%%%%%%%%%%%%%%%%%%%%%%%%%%%%%%%%%%%%%%%%%%%%%%%%%%%%%%%%%%%%%%%%%%%%
\small{
}
%%%%%%%%%%%%%%%%%%%%%%%%%%%%%%%%%%%%%%%%%%%%%%%%%%%%%%%%%%%%%%%%%%%%%%%%
\end{document}